\documentclass{article}

\usepackage{microtype}
\usepackage{graphicx}
\usepackage{subcaption}
\usepackage{booktabs} 

\usepackage{hyperref}

\usepackage[utf8]{inputenc} 
\usepackage[T1]{fontenc}    
\usepackage{hyperref}       
\usepackage{url}            
\usepackage{booktabs}       
\usepackage{amsfonts}       
\usepackage{nicefrac}       
\usepackage{microtype}      
\usepackage{xcolor}         
\usepackage{graphicx}
\usepackage{amsmath}
\usepackage{float} 
\usepackage{amssymb}

\usepackage{adjustbox}
\usepackage{graphicx} 
\usepackage{pifont}  
\usepackage{enumitem}
\usepackage{tabularx}

\usepackage{xcolor}
\usepackage[dvipsnames]{xcolor}

\usepackage[preprint]{icml2026}

\usepackage{amsmath}
\usepackage{amssymb}
\usepackage{mathtools}
\usepackage{amsthm}

\usepackage[capitalize,noabbrev]{cleveref}

\theoremstyle{plain}
\newtheorem{theorem}{Theorem}[section]

\theoremstyle{definition}
\newtheorem{definition}[theorem]{Definition}

\theoremstyle{remark}

\usepackage[textsize=tiny]{todonotes}

\icmltitlerunning{Local Intrinsic Dimension of Representations Predicts Alignment and Generalization}

\begin{document}

\twocolumn[
  \icmltitle{Local Intrinsic Dimension of Representations Predicts Alignment and Generalization in AI Models and Human Brain}



  \icmlsetsymbol{equal}{*}

  \begin{icmlauthorlist}

  
    \icmlauthor{Junjie Yu}{equal,sustech}
    \icmlauthor{Wenxiao Ma}{equal,sustech}
    \icmlauthor{Chen Wei}{sustech}
    \icmlauthor{Jianyu Zhang}{sustech}
    \icmlauthor{Haotian Deng}{sustech}
    \icmlauthor{Zihan Deng}{sustech}
    \icmlauthor{Quanying Liu}{sustech}
  \end{icmlauthorlist}

  \icmlaffiliation{sustech}{Department of Biomedical Engineering, Southern University of Science and Technology}

  \icmlcorrespondingauthor{Quanying Liu}{liuqy@sustech.edu.cn}

  \icmlkeywords{Machine Learning, ICML}

  \vskip 0.3in
]




\printAffiliationsAndNotice{\icmlEqualContribution}

\begin{abstract}
Recent work has found that neural networks with stronger generalization tend to exhibit higher representational alignment with one another across architectures and training paradigms. In this work, we show that models with stronger generalization also align more strongly with human neural activity. Moreover, generalization performance, model--model alignment, and model--brain alignment are all significantly correlated with each other.
We further show that these relationships can be explained by a single geometric property of learned representations: the local intrinsic dimension of embeddings. Lower local dimension is consistently associated with stronger model--model alignment, stronger model--brain alignment, and better generalization, whereas global dimension measures fail to capture these effects. Finally, we find that increasing model capacity and training data scale systematically reduces local intrinsic dimension, providing a geometric account of the benefits of scaling. Together, our results identify local intrinsic dimension as a unifying descriptor of representational convergence in artificial and biological systems.
\end{abstract}



\section{Introduction}

Recent advances in large-scale foundation models have revealed a striking phenomenon: as artificial neural networks scale up in data and capacity, their internal representations exhibit increasing alignment with one another, despite differing architectures and training objectives \citep{huh2024position}. Even more remarkably, these models align with human neural activity, particularly in the visual cortex, even though they are trained without any supervision from biological data \citep{yamins2014performance, khaligh2014deep, gucclu2015deep, eickenberg2017seeing, wang2023better, conwell2024large, shen2024towards, raugel2025disentangling}. This emerging evidence suggests a "Platonic Representation Hypothesis": that both biological and artificial intelligence, when optimized for complex tasks, converge toward a shared, privileged representation of the external world \citep{huh2024position}.

While the existence of representational convergence is increasingly well-documented, prior studies have typically examined AI-AI alignment and AI-Brain alignment separately \citep{muttenthaler2025aligning, mahner2025dimensions, huh2024position}. As a result, it remains unclear whether the representations that spontaneously converge across artificial models are the same ones that align well with human neural activity. Moreover, existing analyses primarily document the existence of alignment, without offering a mathematical or geometric language to describe the convergent representations themselves. This leaves open fundamental questions about what these shared representations look like, what structural properties they possess, and why they are preferentially adopted by both artificial and biological systems.

\begin{figure}[!t]
  \centering
  \includegraphics[width=0.9\linewidth]{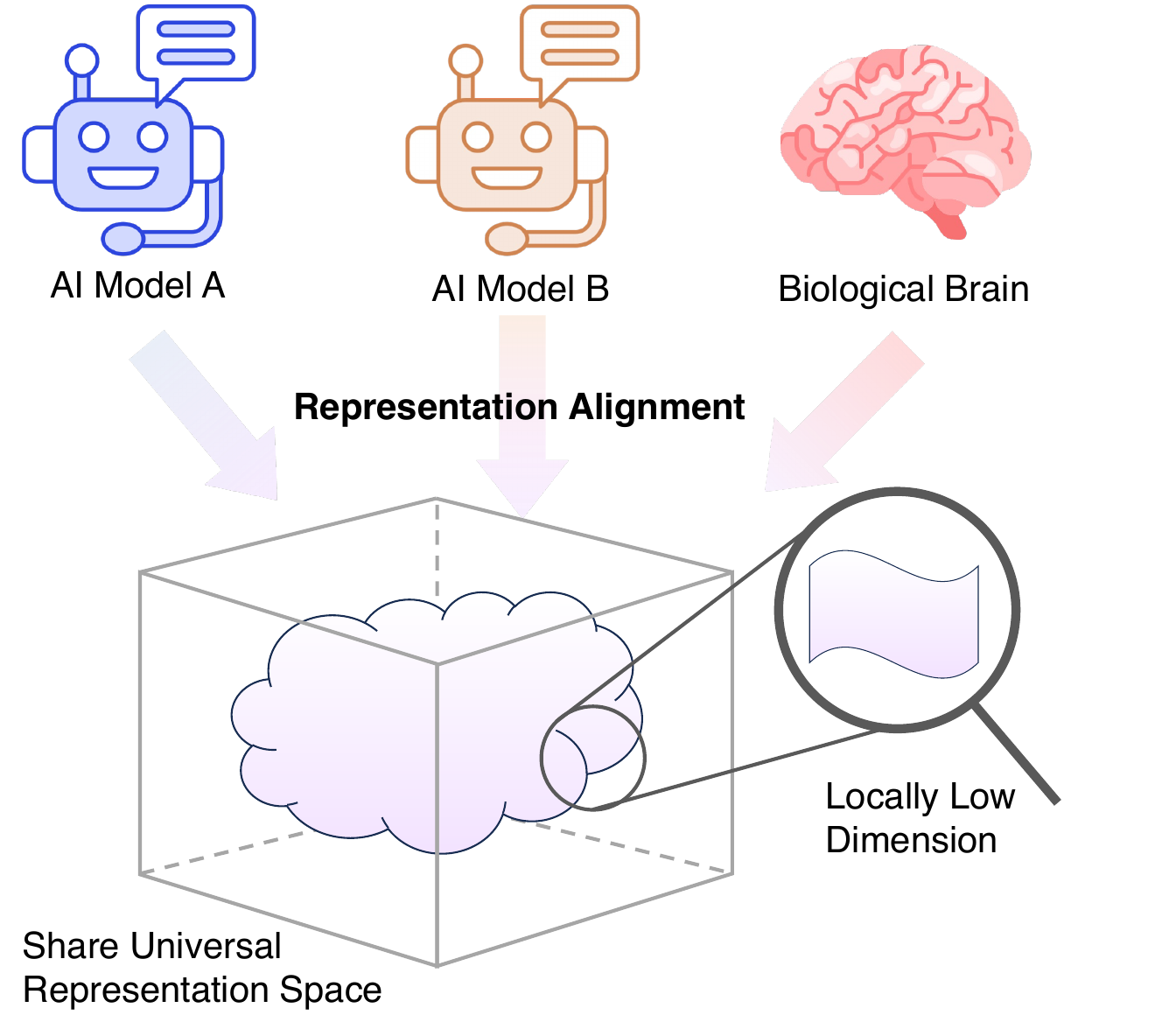}
    \caption{\textbf{Motivation and overview of representational alignment.} 
    Higher-performing AI models exhibit more similar representations to each other and to human neural activity. 
    These convergent representations are also locally low-dimensional, revealing simple geometric principles underlying cross-model and AI-brain alignment.}
  \label{fig:motivation}
\end{figure}

In this work, we conduct a large-scale joint analysis of vision models and human fMRI recordings to bridge this gap. We provide robust empirical evidence that AI-AI alignment, AI-Brain alignment, and generalization are systematically linked and co-vary across models. We find that models with stronger generalization performance tend to exhibit greater representational alignment both with other AI systems and with the human brain. This indicates that the representations favored by both artificial models and the human brain are not arbitrary high-dimensional encodings, but possess a specific and systematic structure that supports effective encoding and generalization.

What geometric properties define these shared representations? Motivated by prior findings linking representational dimensionality to generalization \citep{ansuini2019intrinsic}, we hypothesize that dimension may also be a key factor underlying AI-AI and AI-Brain alignment. Through large-scale experiments, we confirm that representational dimension is significantly associated with both AI-AI and AI-Brain alignment, as well as with generalization performance. Moreover, by conducting a multi-scale analysis, we find that local intrinsic dimension is a more sensitive and informative indicator than global dimensionality (Figure \ref{fig:motivation}).

Finally, we provide a geometric account of scaling laws. We demonstrate that increasing training data and model capacity systematically drives down the local intrinsic dimension of representations. This suggests that the benefits of scaling arise from a progressive refinement of local geometric structure, yielding representations that are increasingly constrained to lower-dimensional neighborhoods in representational space.

Our contributions are threefold:

\begin{itemize}[left=0pt,align=left]
    \item \textbf{Unified Framework}: We establish the strong correlation between AI-AI alignment, AI-Brain alignment, and Generalization.
    \item \textbf{Geometric Principle}: We identify Local Intrinsic Dimensionality as the key geometric descriptor of this convergence.
    \item \textbf{Scaling Mechanism}: We show that increasing model and data scale systematically reduces local dimensionality, providing a geometric explanation for the success of foundation models.
\end{itemize}

\section{Related Work}

\paragraph{AI-Brain Representational Alignment.}
Early research in neuro-AI established that deep neural networks, particularly those trained on object classification, could predict neural responses in the primate visual system \citep{yamins2014performance, khaligh2014deep, gucclu2015deep, eickenberg2017seeing, schrimpf2021neural}. These findings were often interpreted as a consequence of task-specific supervised optimization, with brain alignment viewed as a byproduct of training on category labels. More recent work has shown that such alignment can emerge even in large-scale foundation models trained without any neural supervision, and that alignment improves systematically with model and dataset scale \citep{wang2023better, conwell2024large, shen2024towards, raugel2025disentangling}. However, no existing work can explain why certain models spontaneously align with the brain, or what properties drive this alignment.

\paragraph{Representational Convergence and the Platonic Hypothesis.}
Parallel to model-brain comparison, growing evidence suggests that artificial models themselves are converging toward a universal representational structure. Despite differences in architecture and training objectives, high-performing models increasingly learn similar representations \citep{huh2024position}. This trend has led to the "Platonic Representation Hypothesis", which posits that sufficiently powerful learning systems naturally converge to a shared, optimal statistical model of the underlying reality. In this view, both biological and artificial intelligence are approximations of the same ideal representation.


\paragraph{Intrinsic Dimensionality and Generalization.}
Previous studies have established a link between representational dimensionality and generalization \citep{ansuini2019intrinsic, sharma2022scaling}. However, these analyses have typically relied on almost exclusively local estimates, overlooking the multi-scale nature of representation geometry. Building on this foundation, we perform a comprehensive, multi-scale analysis and find that it is the local intrinsic dimensionality, rather than global estimates, that is most strongly correlated with generalization performance. This highlights that representational "quality" is encoded in the compression of local neighborhoods, providing a more nuanced geometric account of why some models generalize better than others.

\section{Preliminaries and Technical Background}
\label{sec:preliminaries}

This section introduces the datasets, notation and technical components used throughout the paper.
We first describe the experimental dataset and its paired stimulus--response structure (Section~\ref{subsec:Dataset}).
We then outline the AI--fMRI alignment pipeline and the extraction of model embeddings (Section~\ref{subsec: AI-fMRI alignment}), followed by the AI--AI alignment procedure used to compare representations across models (Section~\ref{subsec: AI-AI alignment}).
Finally, we introduce intrinsic dimensionality and describe the estimator employed in our analysis (Section~\ref{subsec:intrinsic_dim}).

\subsection{Dataset Description}
\label{subsec:Dataset}

We use the \emph{Natural Scenes Dataset (NSD)}~\cite{allen2022massive}, which provides high-resolution fMRI responses from human subjects viewing a large set of natural images. Each image stimulus is presented multiple times, allowing for reliable estimation of stimulus-evoked neural activity. The dataset contains thousands of distinct images, covering a wide range of natural scene categories, along with corresponding whole-brain fMRI recordings from each participant. The Natural Scenes Dataset has become a widely adopted benchmark in neuroscience for studying visual representations, and has been used extensively in analyses ranging from encoding and decoding models to representational similarity and brain-model alignment studies \cite{doerig2023neuroconnectionist, takagi2023high, scotti2023reconstructing, doerig2025high, prince2024contrastive}.

This paired stimulus--response design enables a direct, stimulus-level assessment of model--brain alignment. Given a visual stimulus, we can also feed the same image into a vision model to extract internal representations, which are then compared to the neural responses elicited by that image.

\subsection{AI--fMRI Alignment Pipeline}
\label{subsec: AI-fMRI alignment}

For a given pretrained vision model, we first compute embeddings for all images in the dataset, yielding a feature matrix in which each row corresponds to an image and each column to a feature dimension. The images are then split into training and test sets. To control for differences in embedding dimensionality across models, we fit a PCA on the training embeddings and project both training and test embeddings onto the top 300 principal components to avoid data leakage.
For each voxel, we train a ridge regression model to predict its fMRI responses from the model embeddings using the training set. Performance is evaluated on the held-out test set using the coefficient of determination ($R^2$), which we refer to as the \emph{alignment score}. Further details on preprocessing, cross-validation, and hyperparameter selection are provided in Appendix~\ref{supple: fMRI Alignment}.

\subsection{AI-AI alignment pipeline}
\label{subsec: AI-AI alignment}

To assess alignment between different models, we adopt an embedding-based procedure that mirrors the AI--fMRI alignment pipeline. For a shared set of images, embeddings are computed for each model and the images are split into training and test sets. PCA is fit on the training embeddings and used to project both training and test embeddings onto the top 300 principal components, ensuring that dimensionality differences do not bias the regression.
Using ridge regression, we predict one model's embeddings from another's on the training set and evaluate predictive performance on the held-out test set using $R^2$. This $R^2$ serves as the alignment score between models, providing a measure of how similarly different models represent the same input data. Since predicting Model A from Model B and Model B from Model A can yield slightly different $R^2$ values, we compute both directions and take the mean as the final alignment score. By mirroring the AI--fMRI procedure, this approach enables a consistent comparison of representations across both brain and model spaces.

\begin{figure*}[!t]
  \centering
  \includegraphics[width=0.98\linewidth]{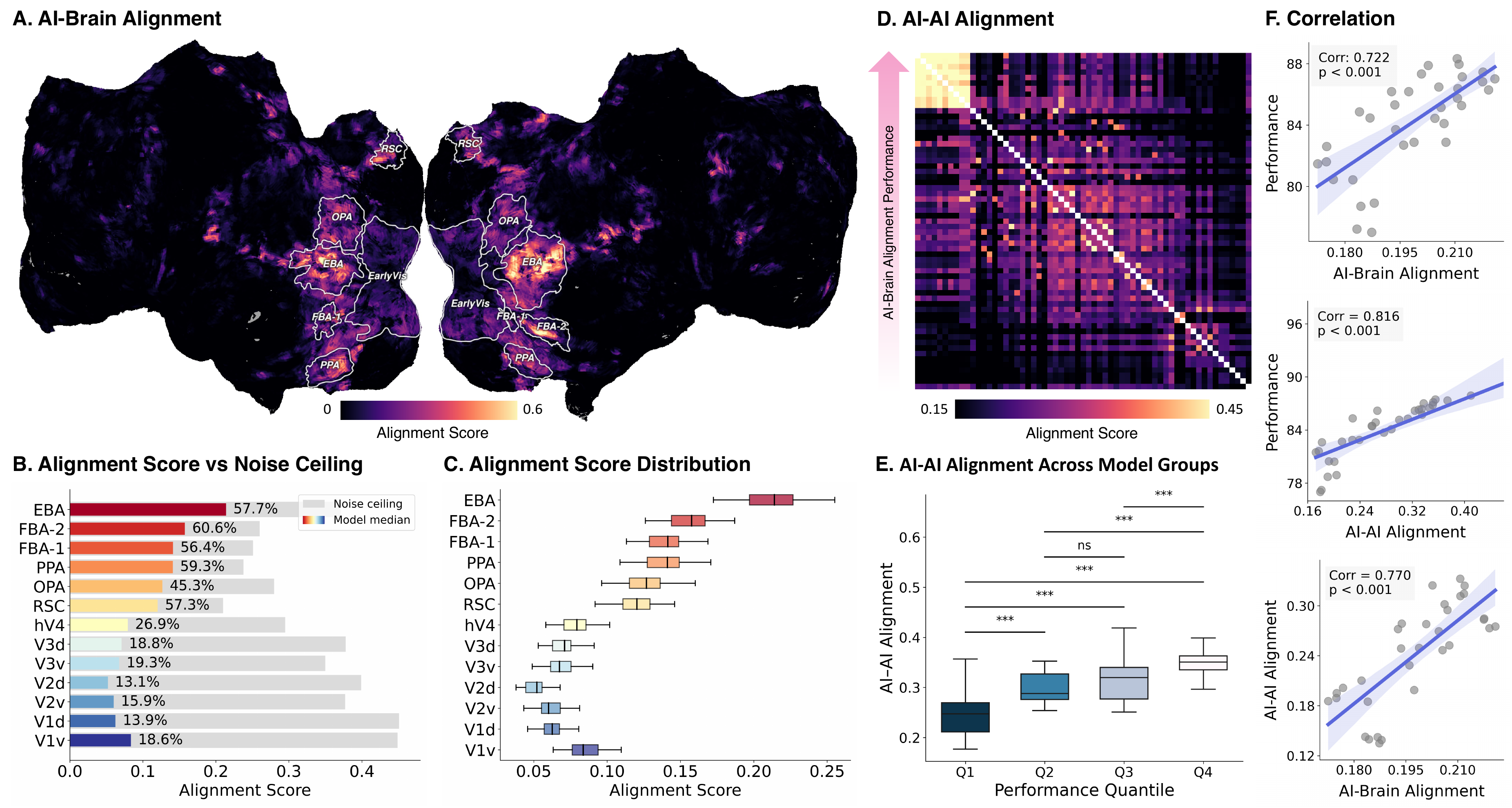}
    \caption{
    \textbf{Representational convergence links AI--Brain alignment, inter-model alignment, and generalization.}
    (\textbf{A}) Whole-brain maps of AI--Brain alignment, showing strongest alignment in visual cortex.
    (\textbf{B}) Median alignment across cortical regions, normalized by noise ceiling; higher-level visual areas (e.g., EBA) reach up to $\sim$60\% of ceiling.
    (\textbf{C}) Distributions of alignment scores across models for each region, revealing substantial inter-model variability.
    (\textbf{D}) Pairwise AI--AI alignment matrix, with models ordered by AI--Brain alignment; models with stronger brain alignment also exhibit stronger mutual alignment.
    (\textbf{E}) Distributions of AI--AI alignment within groups of models binned by ImageNet performance; higher-performing groups exhibit stronger inter-model alignment.
    (\textbf{F}) Pairwise correlations between AI--Brain alignment, AI--AI alignment, and generalization performance, showing that these measures are all positively associated across models.
    }
  \label{fig:representationalConvergence}
\end{figure*}

\subsection{Intrinsic dimensionality and its estimation}
\label{subsec:intrinsic_dim}

Given a set of embeddings 
\(Z = \{z_i\}_{i=1}^N \subset \mathbb{R}^d\), 
we quantify their intrinsic dimensionality, which captures the complexity of the embedding set.

For an anchor point \(z \in Z\), let \(T_j(z)\) denote the Euclidean distance to its \(j\)-th nearest neighbor, and \(\mathcal{N}_K(z)\) the set of its \(K\) nearest neighbors. The local intrinsic dimensionality at \(z\) is estimated as
\[
\hat m_K(z)
=
\left[
\frac{1}{K-1}
\sum_{j=1}^{K-1}
\log\!\left(
\frac{T_K(z)}{T_j(z)}
\right)
\right]^{-1},
\]  
and the scale-\(K\) intrinsic dimensionality is obtained by averaging over all points,
\[
\bar m(K) = \frac{1}{|Z|} \sum_{z \in Z} \hat m_K(z).
\]  

The neighborhood size \(K\) acts as a spatial scale: small \(K\) captures local geometry, whereas larger \(K\) produces a more global estimate of dimensionality. By varying \(K\), \(\bar m(K)\) characterizes representational complexity across scales. Additional details are provided in Appendix \ref{supple: details of dimension}  

In this work, we use the maximum-likelihood estimator (MLE) of Levina and Bickel~\citep{levina2004maximum} to compute \(\hat m_K(z)\). For completeness, we also compared alternative estimators, including the Method Of Moments algorithm (MOM) \citep{amsaleg2018extreme} and Manifold-Adaptive Dimension Estimation algorithm (MADA) \citep{farahmand2007manifold}, in Appendix~\ref{supple:estimator_differences}.

\section{Results}

\subsection{Evidence for representational convergence}
\label{ssec:alignment_convergence}

We first ask whether artificial vision models and the human visual system converge toward a shared representational structure.
To address this question, we analyze representational alignment at three complementary levels: alignment between model representations and fMRI responses (AI--Brain), similarity across model representations (AI--AI), and the relationship of both to generalization performance.
Our goal is to determine whether improvements in generalization are systematically accompanied by convergence in representation space.

\paragraph{AI--Brain alignment across visual cortex.}
We quantify alignment between pretrained vision models and human fMRI responses elicited by the same set of natural images.
As shown in Figure~\ref{fig:representationalConvergence}A, the distribution of alignment scores across the whole brain highlights that the regions with the strongest alignment are predominantly in the visual cortex.
Figure~\ref{fig:representationalConvergence}B summarizes the median alignment score across models for each brain region, alongside estimated noise ceilings derived from repeated fMRI measurements (for details, see Appendix \ref{supple: fMRI Alignment}). The noise ceiling reflects the theoretical maximum Alignment score that could be achieved. In higher-level visual areas such as EBA and FBA, the median alignment of large models reaches roughly 60\% of this ceiling, indicating a high degree of correspondence.
Nonetheless, Figure~\ref{fig:representationalConvergence}C shows substantial variability in alignment across models even within a single region such as EBA, indicating considerable differences between models.

\paragraph{Representational convergence across models.}
We next examine representational similarity across artificial models.
Figure~\ref{fig:representationalConvergence}D shows the pairwise AI--AI alignment matrix, with models ordered by AI--Brain alignment performance.
This ordering reveals a clear pattern: models that align more strongly with the brain also exhibit higher mutual alignment with other models, forming a coherent high-alignment block, whereas lower-performing models show weaker and more heterogeneous alignment.
This observation suggests that brain-aligned models tend to converge with one another, but it remains unclear whether convergence also relates directly to generalization performance.

\paragraph{Convergence increases systematically with generalization.}
To test this directly, we group models according to ImageNet-1K performance and compute average within-group AI--AI alignment.
As shown in Figure~\ref{fig:representationalConvergence}E, within-group alignment increases monotonically with generalization performance.
This confirms that convergence is not driven by models selected for brain alignment, but emerges consistently among models that generalize well.

\paragraph{Reference-based alignment links convergence, brain alignment, and performance.}
To facilitate a quantitative analysis of AI--AI alignment, we define a reference-based measure using the highest-performing model (in terms of generalization) as a reference.
Importantly, this reference is chosen solely based on performance and does not rely on neural data.
By contrast, using the AI--Brain best model as a reference would produce a trivially correlated result between AI--AI and AI--Brain alignment, which we want to avoid.

As shown in Figure~\ref{fig:representationalConvergence}F, reference-based AI--AI alignment is strongly correlated with both AI--Brain alignment and generalization performance.
Models that are closer to the generalization-optimal reference not only generalize better, but also align more strongly with neural responses, confirming that convergence in representation space reflects meaningful performance-related structure rather than trivial correlation.

Together, these analyses provide converging evidence that improved generalization is accompanied by systematic convergence in representation space.
As performance increases, model representations become increasingly similar to one another and increasingly aligned with the human visual cortex, suggesting the emergence of a shared representational structure.

\subsection{Intrinsic dimensionality captures model convergence and generalization}
\label{ssec:dim_characterization}

Having established that AI representations converge with one another and with neural responses, we next ask whether a simple, measurable property of these representations can simultaneously characterize AI--AI alignment, AI--Brain alignment, and generalization performance.
We focus on intrinsic dimensionality as a task-agnostic descriptor of representational geometry.

\begin{figure}[!t]
  \centering
  \includegraphics[width=0.98\linewidth]{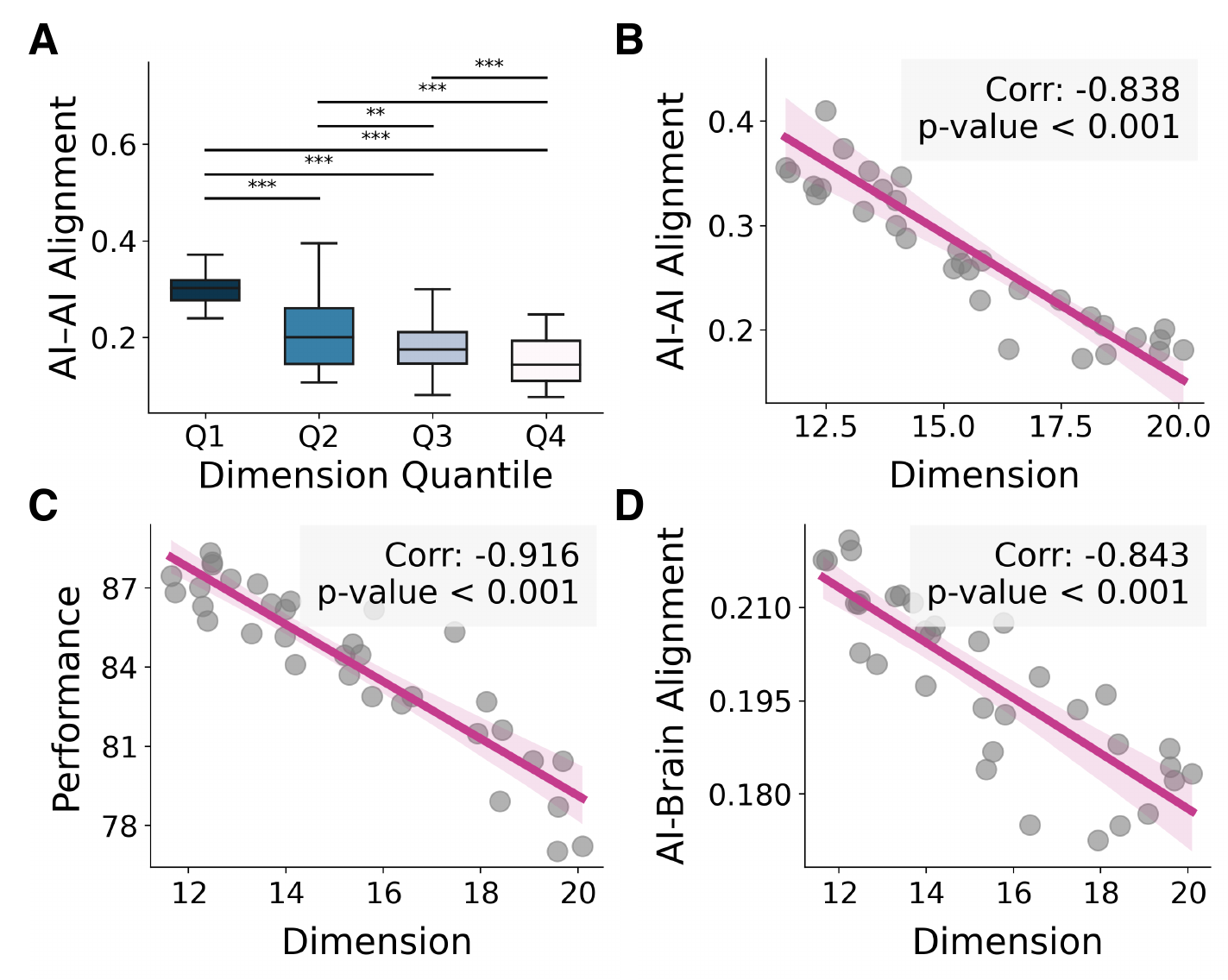}
    \caption{
    \textbf{Intrinsic dimensionality and representational convergence.}
    (\textbf{A}) Models are grouped into bins according to intrinsic dimensionality, and the distribution of within-group AI--AI alignment is shown for each bin; lower-dimensional models align more strongly with each other.
    (\textbf{B}) Intrinsic dimensionality is negatively correlated with AI--AI alignment measured relative to the generalization-optimal reference model.
    (\textbf{C}) Intrinsic dimensionality is negatively correlated with AI--Brain alignment in EBA.
    (\textbf{D}) Intrinsic dimensionality is negatively correlated with ImageNet-1K performance.
    Across all measures, lower intrinsic dimensionality consistently corresponds to stronger representational alignment and improved generalization.
    }
  \label{fig:dimensionResult}
\end{figure}

\paragraph{Dimensionality organizes representational convergence.}
We first group models into four bins according to intrinsic dimensionality and examine AI--AI alignment within each group.
As shown in Figure~\ref{fig:dimensionResult}A, models with lower intrinsic dimensionality exhibit substantially stronger within-group alignment, whereas higher-dimensional models show weaker and more dispersed alignment.
This demonstrates that representational convergence across models is systematically structured by intrinsic dimensionality.

\paragraph{Dimensionality predicts alignment and generalization.}
We next analyze intrinsic dimensionality at the level of individual models.
Figures~\ref{fig:dimensionResult}B--D show that lower intrinsic dimensionality is significantly associated with stronger AI--AI alignment to the reference model, stronger AI--Brain alignment, and higher generalization performance.
All relationships are statistically significant.

\begin{figure*}[!t]
  \centering
  \includegraphics[width=0.95\linewidth]{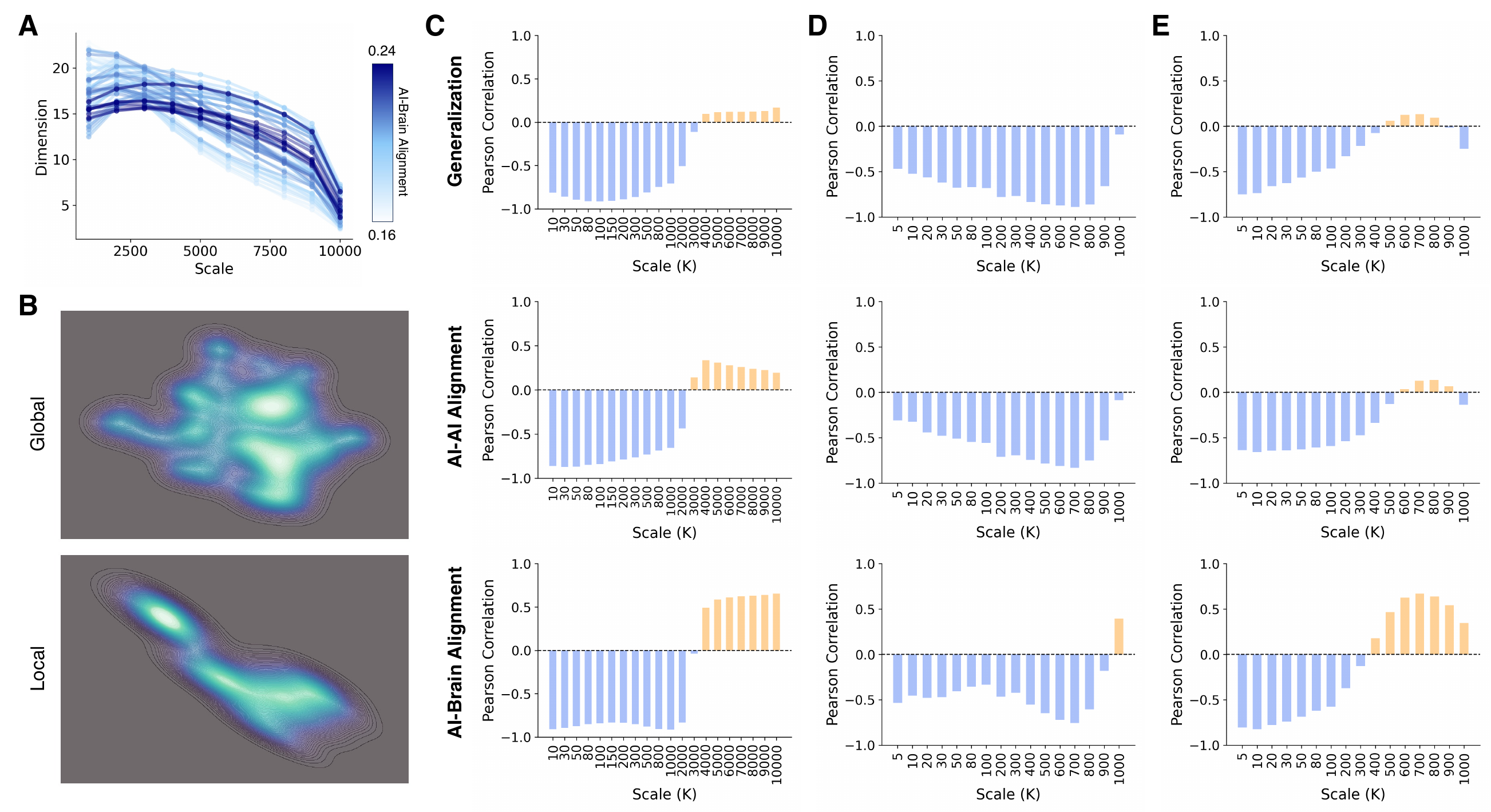}
    \caption{
    \textbf{Alignment and generalization depend on local, not global, intrinsic geometry.}
    (\textbf{A}) Intrinsic dimensionality estimated as a function of neighborhood size $K$, spanning local to global scales.
    (\textbf{B}) Visualization of local versus global embedding structure with matched sample sizes.
    (\textbf{C}) Correlations between intrinsic dimensionality and AI--AI alignment, AI--Brain alignment, and generalization across scales, with strongest effects at local scales.
    (\textbf{D}) Correlations computed within a fixed local neighborhood (1{,}000 nearest neighbors), remaining stable across a range of $K$.
    (\textbf{E}) Control analysis using random global subsampling with matched sample size, demonstrating that the observed effects are not driven by differences in data volume.
    }
  \label{fig:scaleResults}
\end{figure*}

These results suggest that intrinsic dimensionality provides a simple and interpretable measure of representational convergence.
Models with more compressed, lower-dimensional representations not only become more similar to each other, but also align better with neural responses and generalize better, linking geometric simplicity to both biological relevance and functional performance.

\subsection{Alignment depends on local rather than global structure}
\label{ssec:scale_analysis}

We previously analyzed intrinsic dimension at a specific scale (K=100). Here, we extend this analysis across multiple scales to ask whether the relationship between intrinsic dimension, alignment and generalization depends on the spatial scale at which dimensionality is estimated.

\paragraph{Local dimensionality is most predictive.}
We first estimate intrinsic dimensionality across a range of neighborhood sizes (Figure~\ref{fig:scaleResults}A), observing that estimated dimensionality systematically decreases as neighborhood size increases. To further interpret these results, we visualized the embeddings at both global and local scales. For the global distribution, we randomly sampled 1,000 embeddings, while for the local distribution, we selected an anchor point and extracted its 1,000 nearest neighbors. Using t-SNE, we find that the global embeddings contain more distinct clusters with greater heterogeneity in inter-point distances, a hallmark of low-dimensional structure, whereas the local neighborhoods also form clusters, but fewer and more evenly distributed, consistent with higher local dimensionality (Figure~\ref{fig:scaleResults}B). These visualizations corroborate our multi-scale analysis and highlight the distinct geometric properties captured at different scales.

Next, in Figure~\ref{fig:scaleResults}C, we compute correlations between dimensionality at different scales and three measures.
We find that dimensionality estimated at small (local) scales is most strongly predictive, whereas correlations weaken as the neighborhood size increases, indicating that alignment- and performance-relevant structure is more closely associated with local geometry.

\paragraph{Local geometry, not sample size, drives the effect.}
To ensure that these effects are due to local structure rather than estimation hyperparameters, we fix the neighborhood size and compute dimensionality within locally sampled neighborhoods.
Specifically, for each model, we randomly select a reference sample and extract its 1,000 nearest neighbors in embedding space to estimate local dimensionality.
As shown in Figure~\ref{fig:scaleResults}D, the resulting dimensionality remains significantly negatively correlated with generalization and alignment metrics across neighborhood sizes, confirming that local geometric structure, rather than hyperparameters, governs these relationships.
For comparison, we perform the same analysis on subsets obtained by randomly sampling 1,000 embeddings from the entire dataset (Figure~\ref{fig:scaleResults}E).
In this case, correlations weaken and show scale-dependent effects, demonstrating that the predictive power of dimensionality arises specifically from local embedding structure, not from sample count or estimation procedure.

Together, these analyses show that representational alignment and generalization are primarily shaped by local, rather than global, geometry.

\subsection{Architecture-dependent structure of representational alignment}
\label{ssec:cross_architecture}

The analyses above focused on ConvNeXt models.
We next extend these observations to a broader set of architectures, ViT (3 models), ConvNeXt (20 models), ResNet (20 models), and ResMLP (8 models), to examine whether architectural differences substantially affect representational convergence, brain alignment, and generalization. 
To reduce the influence of hyperparameter choices, we selected the optimal estimation scale for each analysis based on the results from the previous section (AI-AI alignment and generalization used K=50, AI-Brain alignment used K=1000).
Details of the models are provided in the Appendix \ref{supple:model_details}.

\begin{figure}[!t]
  \centering
  \includegraphics[width=0.98\linewidth]{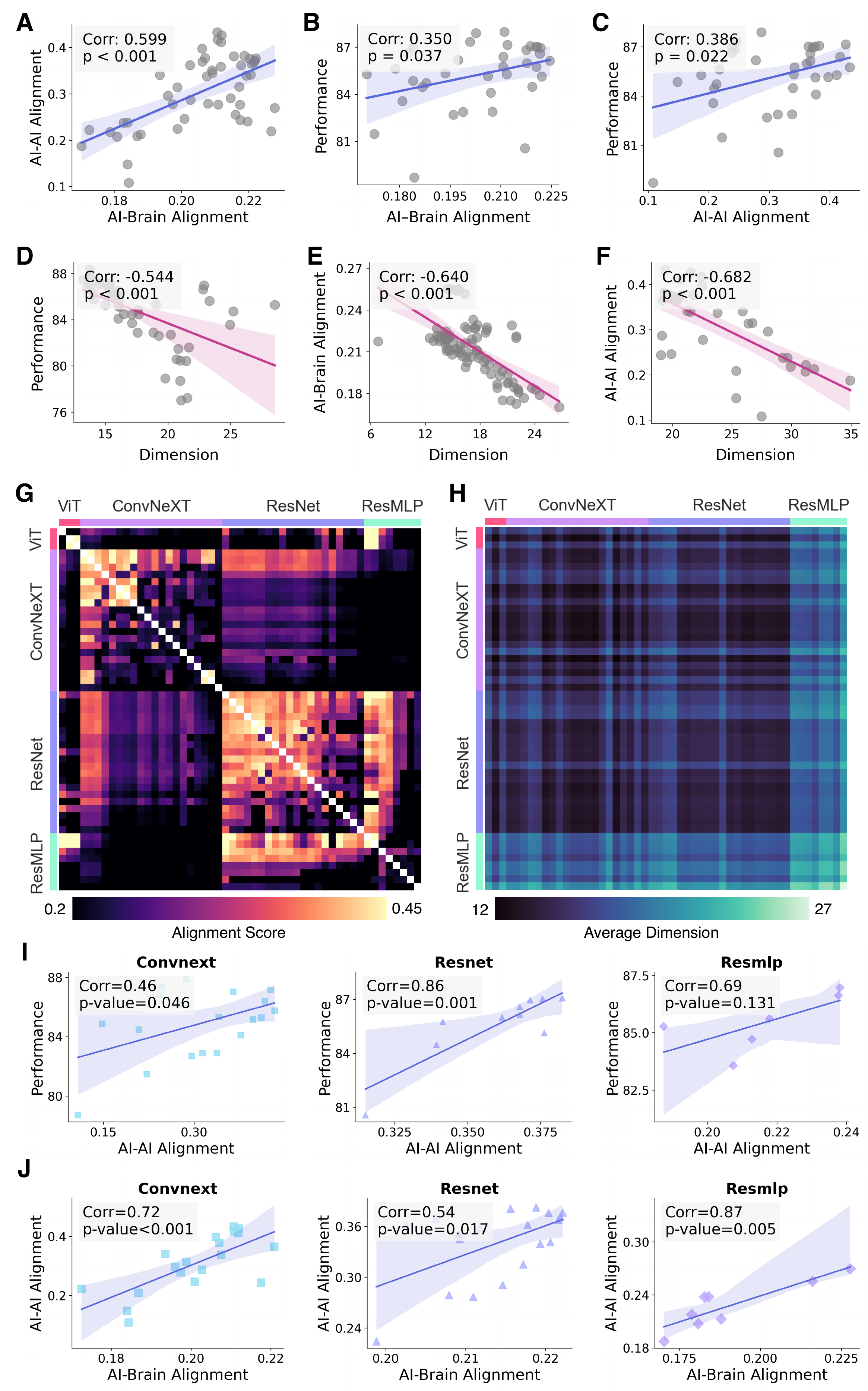}
    \caption{
    \textbf{Representational alignment generalizes across architectures.}
    (\textbf{A--C}) Relationships between AI--AI alignment (measured relative to the generalization-optimal reference), AI--Brain alignment, and ImageNet-1K performance when models from all architectures are pooled.
    (\textbf{D--F}) Relationships between intrinsic dimensionality and AI--AI alignment, AI--Brain alignment, and generalization across architectures.
    (\textbf{G}) Pairwise AI--AI alignment matrix reveals family-level structure, with stronger alignment within architectural families than across families.
    (\textbf{H}) Average intrinsic dimensionality shows no systematic differences across architectures.
    (\textbf{I}) Correlations between generalization performance and AI--AI alignment across models from different architectures.
    (\textbf{J}) Correlations between AI--AI alignment and AI--Brain alignment across architectures.
    }
  \label{fig:multiModel}
\end{figure}

\paragraph{Alignment and dimensionality correlations persist across architectures.}
Figures~\ref{fig:multiModel}A--F summarize relationships between AI--AI alignment (relative to the generalization-optimal reference), AI--Brain alignment, generalization performance, and intrinsic dimensionality across all models.
As expected, combining models from different architectures slightly reduces the strength of correlations compared with within-architecture analyses. Nevertheless, the overall relationships remain statistically significant, indicating that the fundamental links between dimensionality, alignment, and performance are robust across heterogeneous model architectures.

\paragraph{Architectural family shapes inter-model alignment.}
The full pairwise AI--AI alignment matrix (Figure~\ref{fig:multiModel}G) reveals that inter-model alignment is strongly influenced by architectural family. Models within the same architecture, such as ConvNeXt or ResNet, form distinct clusters with notably higher alignment, whereas alignment between models from different architectures is substantially lower. Importantly, average intrinsic dimensionality does not differ systematically across architectures (Figure~\ref{fig:multiModel}H), indicating that the reduced cross-architecture alignment cannot be explained by dimensionality differences alone.

\paragraph{Cross-architecture convergence emerges at high performance.}
Interestingly, even models with very different architectures show significant correlations between their alignment to the generalization-optimal reference and both generalization performance and AI--Brain alignment (Figure~\ref{fig:multiModel}I--J). 
In our dataset, the highest-performing model, \texttt{convnext\_large\_mlp.clip\_laion2b \_soup\_ft\_in12k\_in1k\_384} from the ConvNeXt family, serves as this reference. 
Models from other architectural families, such as ResNet, exhibit alignment patterns that are significantly correlated with this top-performing model. 
These results indicate that, although architectural biases introduce noticeable differences at low- and mid-level representations, models with sufficiently high performance tend to converge toward similar high-level representations, partially bridging architectural differences.

Together, these findings show that intrinsic dimensionality provides a robust, architecture-agnostic descriptor of representational simplicity, brain alignment, and generalization, whereas AI--AI alignment is strongly shaped by architectural family. 
More detailed analyses within single architectures are presented in the Appendix \ref{supple:architecture_differences}.

\begin{figure}[!t]
  \centering
  \includegraphics[width=0.95\linewidth]{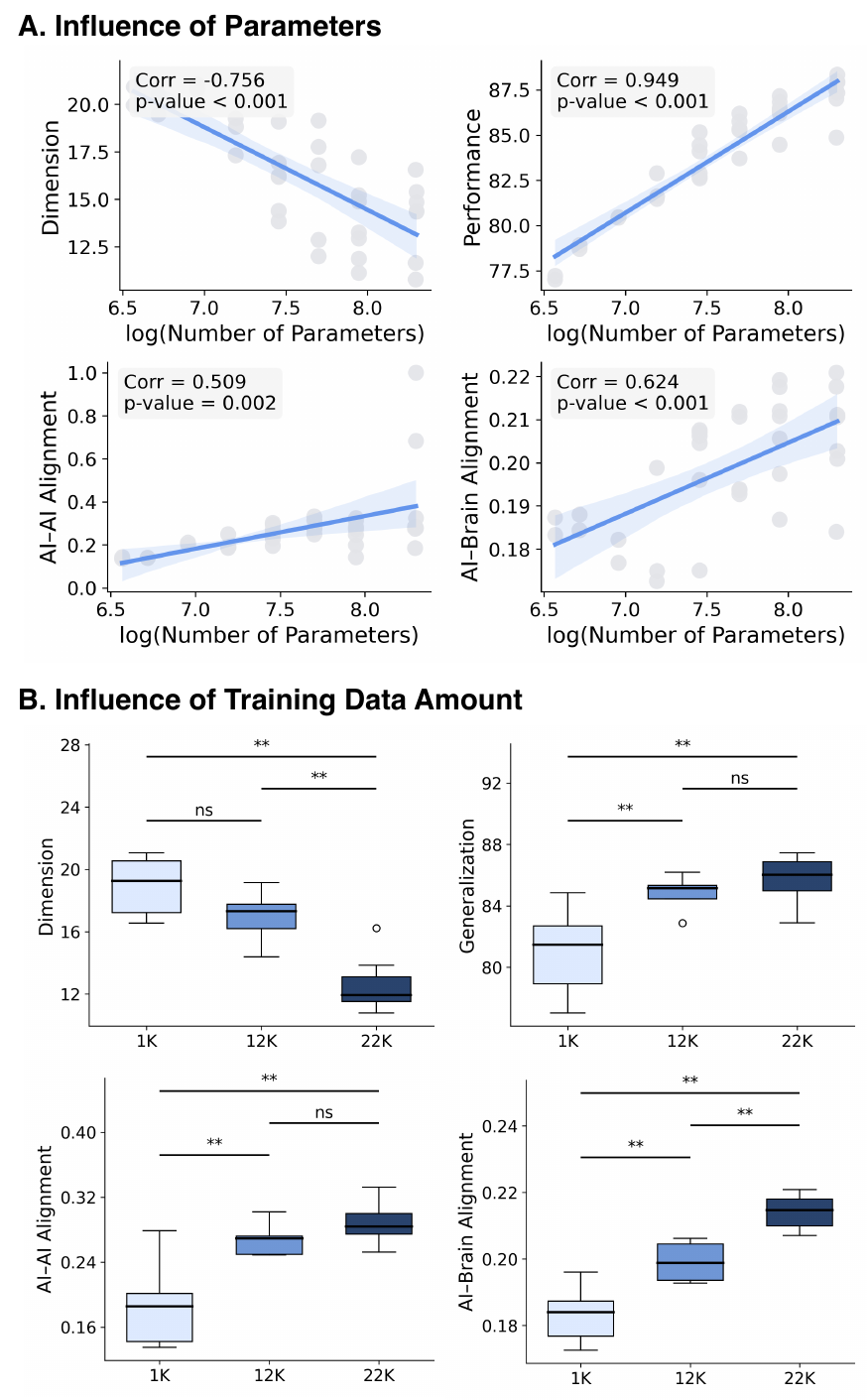}
    \caption{
    \textbf{Training scale modulates local intrinsic dimensionality.}
    (\textbf{A}) Within each architecture, larger models show lower intrinsic dimensionality, stronger AI--AI and AI--Brain alignment, and higher generalization.
    (\textbf{B}) Models trained on larger datasets exhibit decreased local dimensionality alongside improved alignment and performance.
    }
  \label{fig:trainFactors}
\end{figure}

\subsection{Training scale modulates local intrinsic dimensionality}
\label{ssec:training_scale}


Prior work shows that larger models and datasets improve generalization and neural alignment \citep{conwell2024large}. We asked whether these gains are accompanied by systematic changes in local representational geometry.

We first examine the effect of model size and find that increasing the number of parameters is associated with reduced local intrinsic dimensionality, along with improved alignment and generalization performance (Figure~\ref{fig:trainFactors}A).

Because most ConvNext were pretrained on individual datasets (ImageNet-1k, ImageNet-12k, or ImageNet-22k), we can separately assess the impact of dataset size. We find that larger pretraining datasets systematically improve both alignment and generalization while simultaneously reducing local intrinsic dimensionality (Figure~\ref{fig:trainFactors}B).
Overall, increasing model scale or dataset size promotes locally lower-dimensional representations, providing a geometric account for their enhanced neural alignment and generalization.

\section{Discussion}

This work investigates representational convergence in modern vision models through the lens of intrinsic geometry.
By jointly analyzing inter-model alignment, alignment with human neural responses, and generalization performance, we provide the first systematic evidence that higher-performing AI models exhibit increasingly similar representations both to each other and to neural representations in the brain.
Crucially, these convergent representations are characterized by low local intrinsic dimensionality, revealing a simple geometric structure underlying alignment.

We note that all AI--Brain alignment analyses reported in the main text are robust across individual subjects and brain regions.  
Specifically, Appendix~\ref{supple: Subject Differences} presents results for different subjects, while Appendix~\ref{supple: Region Differences} provides analyses across regions.  
These supplementary analyses confirm that the patterns of alignment observed in the main text are consistent and not driven by a subset of subjects or regions.

We interpret the observed relationship between representational alignment and generalization through the lens of flat minima.
A substantial body of prior work has shown that models converging to flatter regions of the loss landscape tend to generalize better \citep{hochreiter1997flat}.
Such flat regions are characterized by the presence of many distinct parameter configurations that achieve similarly high performance.
Although these different parameter settings may induce non-identical embeddings, they correspond to nearby functional solutions and thus give rise to representations that remain mutually alignable.
From this perspective, the stronger representational alignment observed among higher-performing models is consistent with the idea that these models are more likely to reside within the same flat region of the loss landscape.
This view also helps explain why alignment is particularly pronounced within a single architecture, where shared parameter symmetries and optimization landscapes increase the likelihood of convergence to common flat regions \citep{zhao2025symmetry}.

\paragraph{Limitations and Future Work.}
Our analyses focus on vision models and visual cortex responses.
Extending this framework to other modalities, such as language or audition, will be important for assessing the generality of representational convergence.
Moreover, while local intrinsic dimensionality strongly predicts alignment and generalization, the theoretical mechanisms underlying its relevance remain unclear.

\paragraph{Conclusion.}
We present a representation-centric account of convergence in modern vision models grounded in intrinsic geometry.
Local intrinsic dimensionality emerges as a simple, architecture-agnostic link between generalization, inter-model similarity, and brain alignment.
Together, these results suggest that representational convergence reflects the emergence of locally simple, biologically relevant structures shaped by learning and scale.

\section*{Impact Statement}
This paper presents work whose goal is to advance the field of machine learning. 
Our study investigates representational alignment between AI models and the human brain, providing insights that may improve model interpretability and generalization.  
There are many potential societal consequences of this work, none of which we feel must be specifically highlighted here.

\bibliography{example_paper}
\bibliographystyle{icml2026}

\newpage
\appendix
\onecolumn

\section{Overview of Supplementary Analyses}
\label{supple:overview}

This appendix provides a comprehensive set of analyses that extend and validate the main findings of the study. While the main text focuses on EBA and ConvNeXt embeddings, the supplementary analyses explore multiple aspects to ensure the robustness and generality of our conclusions:

\begin{itemize}
    \item \textbf{fMRI Alignment Details (Section~\ref{supple: fMRI Alignment})}: Detailed description of voxel-wise encoding models, evaluation metrics, and noise ceiling estimation for transparency and reproducibility.
    
    \item \textbf{Model Architectures and Implementation (Section~\ref{supple:model_details})}: Examination of multiple architectures (ConvNeXt, ResNet, ResMLP, ViT) with various pretraining regimes to test cross-model generality.
    
    \item \textbf{Dimensional Analysis (Section~\ref{supple: details of dimension})}: Methods for estimating intrinsic dimensionality, including fractal and correlation dimensions, with multi-scale analysis to capture geometric structure of model representations.
    
    \item \textbf{Subject Differences (Section~\ref{supple: Subject Differences})}: Assessment of inter-subject variability in AI-Brain alignment, its correlation with generalization, and dimensionality to ensure findings are not driven by individual participants.
    
    \item \textbf{Region Differences (Section~\ref{supple: Region Differences})}: Evaluation across early, intermediate, and higher-level visual areas to verify region-general and hierarchical patterns in AI-Brain alignment.
    
    \item \textbf{Dimensionality Estimator Robustness (Section~\ref{supple:estimator_differences})}: Comparison of multiple estimators (MLE, MOM, MADA) to confirm scale-dependent dimensionality effects are robust to methodological choices.
\end{itemize}

Collectively, these analyses validate the reproducibility and generality of our findings, demonstrating that AI-Brain alignment reliably reflects representational structure and predicts model generalization across participants, brain regions, architectures, and estimation methods.

\section{Details of fMRI Alignment}
\label{supple: fMRI Alignment}

\paragraph{Voxel-wise encoding model.} 
Let $z_i \in \mathbb{R}^{300}$ be the embedding of image $i$, and let $r_{ij}$ be the fMRI response of voxel $j$ to image $i$. We train a voxel-specific ridge regression model:

\begin{equation}
r_{ij} \approx \mathbf{w}_j^\top \mathbf{z}_i + \epsilon.
\end{equation}

Model fitting is done using 5-fold cross-validation on an 80/20 training/test split. The regularization parameter is tuned via nested cross-validation.

\paragraph{Evaluation metrics.} 
Alignment quality is quantified by the coefficient of determination ($R^2$) on held-out data.

\paragraph{Noise ceiling estimation.}

Due to inherent measurement errors and individual variability in neural signals, even a perfectly accurate model cannot account for all sources of variation. Noise ceiling estimates are derived from the reliability of the beta weights across trials. In essence, the more consistent the neural response is across repeated presentations of the same image, the greater the proportion of the response variance that can be attributed to stimulus-driven signals. These estimates of the noise ceiling establish an upper bound on the amount of variance that can be explained or predicted in the response of a given voxel \citep{allen2022massive}.

To quantify the noise ceiling in a more precise manner, the effective noise variance \( N_{\text{eff}} \) is first computed by considering the number of times each image was presented to the subject, with different weights assigned to each trial type. Let \( A \), \( B \), and \( C \) denote the number of distinct images presented three, two, and one times to the subject, respectively. The effective noise variance \( N_{\text{eff}} \) is calculated as:

\[
N_{\text{eff}} = \frac{\frac{A}{3} + \frac{B}{2} + \frac{C}{1}}{A + B + C}
\]

Subsequently, the noise ceiling \( N_C \) is determined by combining the signal variance \( S^2 \) with the effective noise variance \( N_{\text{eff}} \), as follows:

\[
N_C = \frac{S^2}{S^2 + N_{\text{eff}}}
\]

Here, \( S^2 \) represents the signal variance, which is computed based on the beta weights derived from all NSD scan sessions. The data used is provided by the NSD dataset.

\section{Model Architectures and Implementation Details}
\label{supple:model_details}

In this study, we analyzed multiple convolutional and transformer-based architectures to examine the alignment between neural representations and model embeddings. Our analysis primarily focuses on the ConvNeXt family, while ResNet, ResMLP, and ViT models serve as cross-architecture controls to validate the generality of our findings.

\subsection{ConvNeXt: Primary Analysis Backbone}
\label{supple:convnext}

We employed the ConvNeXt architecture \citep{liu2022convnet} as the main backbone for generating embeddings. ConvNeXt adapts the ResNet architecture with design principles inspired by Vision Transformers, including large kernel sizes, depthwise convolutions, and layer normalization. Its architecture consists of four stages with multiple residual blocks per stage. Key modifications include:

\begin{itemize}
    \item \textbf{GELU activation}: replaces ReLU for smoother gradient propagation.
    \item \textbf{Large convolutional kernels (e.g., 7x7)}: increases the receptive field.
    \item \textbf{Depthwise separable convolutions}: efficient spatial mixing while reducing parameters.
    \item \textbf{Layer normalization instead of batch normalization}: stabilizes training across different datasets.
    \item \textbf{Patchify stem}: aligns preprocessing with Vision Transformer-style models.
\end{itemize}

\paragraph{Advantages of ConvNeXt for controlled analysis:}  
\begin{enumerate}
    \item \textbf{Controlled pretraining}: Each model is pre-trained on a single dataset, allowing systematic comparisons of pretraining effects on brain-model alignment.
    \item \textbf{Architectural similarity}: All models share the same backbone, minimizing confounding effects from structural differences.
    \item \textbf{Scalability across model sizes}: ConvNeXt variants range from \texttt{nano} to \texttt{xlarge}, enabling analyses across scales while keeping architectural principles consistent.
    \item \textbf{Computational efficiency}: Its design allows us to extract embeddings for multiple participants and brain regions without excessive hardware requirements.
\end{enumerate}

\paragraph{Implementation Details:}  
We utilized the Huggingface Transformers API to access ConvNeXt models pre-trained on ImageNet-1K, ImageNet-12K, ImageNet-22K, and LAION datasets. Embeddings were extracted on an NVIDIA RTX 3080 GPU with 10GB memory, sufficient for all model variants. Table~\ref{model-info-table-vision} and Table~\ref{model-info-table-vl} list all ConvNeXt models, including pretraining datasets and whether fine-tuning was applied.

\paragraph{Computational Resources}  
We utilized an NVIDIA GeForce RTX 3080 GPU with 10GB GDDR6X memory, a 320-bit memory interface, and a memory bandwidth of 760 GB/s for embedding extraction using the Huggingface Transformers API. This setup adequately meets the computational requirements without requiring additional resources.



\begin{table}[!t]
  \caption{Vision-only ConvNeXt models without language supervision (Lang = ✗), sorted by EBA alignment score.}
  \label{model-info-table-vision}
  \centering
  \small
  \setlength{\tabcolsep}{4pt}
  \renewcommand{\arraystretch}{1.05}
  \begin{tabular}{lll}
    \toprule
    \textbf{Model Name} & \textbf{Pretraining Dataset} & \textbf{FT} \\
    \midrule
    convnext\_nano.d1h\_in1k & ImageNet-1K & \ding{55} \\
    convnext\_nano\_ols.d1h\_in1k & ImageNet-1K & \ding{55} \\
    convnext\_tiny\_hnf.a2h\_in1k & ImageNet-1K & \ding{55} \\
    convnext\_pico\_ols.d1\_in1k & ImageNet-1K & \ding{55} \\
    convnext\_pico.d1\_in1k & ImageNet-1K & \ding{55} \\
    convnext\_atto\_ols.a2\_in1k & ImageNet-1K & \ding{55} \\
    convnext\_large.fb\_in1k & ImageNet-1K & \ding{55} \\
    convnext\_femto.d1\_in1k & ImageNet-1K & \ding{55} \\
    convnext\_base.fb\_in1k & ImageNet-1K & \ding{55} \\
    convnext\_atto.d2\_in1k & ImageNet-1K & \ding{55} \\
    convnext\_femto\_ols.d1\_in1k & ImageNet-1K & \ding{55} \\

    convnext\_small.in12k\_ft\_in1k\_384 & ImageNet-12K & \ding{51} \\
    convnext\_small.in12k\_ft\_in1k & ImageNet-12K & \ding{51} \\
    convnext\_small.fb\_in1k & ImageNet-1K & \ding{55} \\
    convnext\_tiny.fb\_in1k & ImageNet-1K & \ding{55} \\
    convnext\_nano.in12k\_ft\_in1k & ImageNet-12K & \ding{51} \\
    convnext\_tiny.in12k\_ft\_in1k & ImageNet-12K & \ding{51} \\
    convnext\_tiny.in12k\_ft\_in1k\_384 & ImageNet-12K & \ding{51} \\
    
    convnext\_tiny.fb\_in22k\_ft\_in1k\_384 & ImageNet-22K & \ding{51} \\
    convnext\_tiny.fb\_in22k\_ft\_in1k & ImageNet-22K & \ding{51} \\
    convnext\_small.fb\_in22k\_ft\_in1k\_384 & ImageNet-22K & \ding{51} \\
    convnext\_small.fb\_in22k\_ft\_in1k & ImageNet-22K & \ding{51} \\
    convnext\_small.in12k & ImageNet-12K & \ding{55} \\
    convnext\_base.fb\_in22k\_ft\_in1k\_384 & ImageNet-22K & \ding{51} \\
    convnext\_large.fb\_in22k\_ft\_in1k\_384 & ImageNet-22K & \ding{51} \\
    convnext\_base.fb\_in22k\_ft\_in1k & ImageNet-22K & \ding{51} \\
    convnext\_xlarge.fb\_in22k\_ft\_in1k\_384 & ImageNet-22K & \ding{51} \\
    convnext\_tiny.in12k & ImageNet-12K & \ding{55} \\
    convnext\_large.fb\_in22k\_ft\_in1k & ImageNet-22K & \ding{51} \\
    convnext\_xlarge.fb\_in22k\_ft\_in1k & ImageNet-22K & \ding{51} \\
    convnext\_nano.in12k & ImageNet-12K & \ding{55} \\
    convnext\_small.fb\_in22k & ImageNet-22K & \ding{55} \\
    convnext\_large.fb\_in22k & ImageNet-22K & \ding{55} \\
    convnext\_tiny.fb\_in22k & ImageNet-22K & \ding{55} \\
    convnext\_xlarge.fb\_in22k & ImageNet-22K & \ding{55} \\
    convnext\_base.fb\_in22k & ImageNet-22K & \ding{55} \\
    \bottomrule
  \end{tabular}
\end{table}

\begin{table}[!t]
  \caption{Vision--language ConvNeXt models with language supervision (Lang = ✓), sorted by EBA alignment score.}
  \label{model-info-table-vl}
  \centering
  \small
  \setlength{\tabcolsep}{4pt}
  \renewcommand{\arraystretch}{1.05}
  \begin{tabular}{lll}
    \toprule
    \textbf{Model Name} & \textbf{Pretraining Dataset} & \textbf{FT} \\
    \midrule
    convnext\_base.clip\_laion2b\_augreg\_ft\_in1k & LAION-2B & \ding{51} \\
    convnext\_large\_mlp.clip\_laion2b\_augreg\_ft\_in1k & LAION-2B & \ding{51} \\
    convnext\_large\_mlp.clip\_laion2b\_augreg\_ft\_in1k\_384 & LAION-2B & \ding{51} \\
    convnext\_base.clip\_laiona\_augreg\_ft\_in1k\_384 & LAION-A & \ding{51} \\
    convnext\_large\_mlp.clip\_laion2b\_soup\_ft\_in12k\_in1k\_384 & LAION-2B & \ding{51} \\
    convnext\_base.clip\_laion2b\_augreg\_ft\_in12k\_in1k & LAION-2B & \ding{51} \\
    convnext\_large\_mlp.clip\_laion2b\_soup\_ft\_in12k\_in1k\_320 & LAION-2B & \ding{51} \\
    convnext\_base.clip\_laion2b\_augreg\_ft\_in12k\_in1k\_384 & LAION-2B & \ding{51} \\
    convnext\_xxlarge.clip\_laion2b\_soup\_ft\_in1k & LAION-2B & \ding{51} \\
    convnext\_large\_mlp.clip\_laion2b\_soup\_ft\_in12k\_384 & LAION-2B & \ding{51} \\
    convnext\_large\_mlp.clip\_laion2b\_augreg\_ft\_in12k\_384 & LAION-2B & \ding{51} \\
    convnext\_base.clip\_laion2b\_augreg\_ft\_in12k & LAION-2B & \ding{51} \\
    convnext\_xxlarge.clip\_laion2b\_soup\_ft\_in12k & LAION-2B & \ding{51} \\
    convnext\_large\_mlp.clip\_laion2b\_soup\_ft\_in12k\_320 & LAION-2B & \ding{51} \\
    convnext\_xxlarge.clip\_laion2b\_rewind & LAION-2B & \ding{55} \\
    convnext\_xxlarge.clip\_laion2b\_soup & LAION-2B & \ding{55} \\
    convnext\_large\_mlp.clip\_laion2b\_augreg & LAION-2B & \ding{55} \\
    convnext\_large\_mlp.clip\_laion2b\_ft\_320 & LAION-2B & \ding{51} \\
    convnext\_base.clip\_laiona & LAION-A & \ding{55} \\
    convnext\_large\_mlp.clip\_laion2b\_ft\_soup\_320 & LAION-2B & \ding{51} \\
    convnext\_base.clip\_laion2b & LAION-2B & \ding{55} \\
    convnext\_base.clip\_laiona\_320 & LAION-A & \ding{55} \\
    convnext\_base.clip\_laion2b\_augreg & LAION-2B & \ding{55} \\
    convnext\_base.clip\_laiona\_augreg\_320 & LAION-A & \ding{55} \\
    \bottomrule
  \end{tabular}
\end{table}

\subsection{Cross-Architecture Validation Models}
\label{supple:cross_arch}

To ensure that our findings from ConvNeXt are not architecture-specific, we also evaluated several additional families of models: ResNet, ResMLP, and Vision Transformers (ViT). These models allow us to test whether trends observed in ConvNeXt embeddings generalize across architectural paradigms.

\subsubsection{ResNet Models}
\label{supple:resnet}

We examined ResNet models across different depths (18, 50, 101, 152 layers) and training paradigms (e.g., ImageNet-1K, supervised, self-supervised). The residual structure and hierarchical feature extraction of ResNets provide a natural comparison to ConvNeXt. Table~\ref{tab:resnetsummary} summarizes brain-region alignment results for selected ResNet variants.

\begin{table}[!t]
\caption{Summary of ResNet models across depth and training variants.}
\label{tab:resnetsummary}
\centering
\small
\setlength{\tabcolsep}{5pt}
\renewcommand{\arraystretch}{1.15}
\begin{tabular}{lccc}
\toprule
Model & $R^2_{\text{EBA}}$ & $R^2_{\text{OPA}}$ & $R^2_{\text{PPA}}$ \\
\midrule
resnet18.a1\_in1k                  & 0.211 & 0.131 & 0.150 \\
resnet18.a2\_in1k                  & 0.208 & 0.126 & 0.148 \\
resnet18.a3\_in1k                  & 0.199 & 0.122 & 0.142 \\
resnet18.fb\_ssl\_yfcc100m\_ft\_in1k & 0.215 & 0.133 & 0.152 \\
resnet18.fb\_swsl\_ig1b\_ft\_in1k    & 0.217 & 0.134 & 0.153 \\

resnet50.a1\_in1k                  & 0.222 & 0.135 & 0.154 \\
resnet50.a1h\_in1k                 & 0.209 & 0.126 & 0.144 \\
resnet50.a2\_in1k                  & 0.225 & 0.140 & 0.159 \\
resnet50.a3\_in1k                  & 0.219 & 0.138 & 0.156 \\
resnet50.am\_in1k                  & 0.222 & 0.142 & 0.160 \\

resnet101.a1\_in1k                 & 0.219 & 0.134 & 0.150 \\
resnet101.a1h\_in1k                & 0.206 & 0.121 & 0.140 \\
resnet101.a2\_in1k                 & 0.222 & 0.137 & 0.155 \\
resnet101.a3\_in1k                 & 0.221 & 0.136 & 0.154 \\
resnet101.gluon\_in1k              & 0.214 & 0.130 & 0.148 \\

resnet152.a1\_in1k                 & 0.216 & 0.130 & 0.149 \\
resnet152.a1h\_in1k                & 0.203 & 0.118 & 0.135 \\
resnet152.a2\_in1k                 & 0.220 & 0.136 & 0.152 \\
resnet152.a3\_in1k                 & 0.218 & 0.136 & 0.154 \\
resnet152.gluon\_in1k              & 0.209 & 0.126 & 0.144 \\
\bottomrule
\end{tabular}
\end{table}

\subsubsection{ResMLP Models}
\label{supple:resmlp}

ResMLP is a pure MLP-based vision architecture that replaces convolutions with linear projections while retaining spatial information. We evaluated ResMLP models of varying depths (12, 24, 36 layers) and training regimes. Table~\ref{tab:resmlp-summary} presents the alignment scores for EBA, OPA, and PPA, demonstrating that the general relationship between pretraining and brain alignment extends to MLP-based architectures.

\begin{table}[!t]
\caption{Summary of ResMLP models.}
\label{tab:resmlp-summary}
\centering
\small
\setlength{\tabcolsep}{6pt}
\renewcommand{\arraystretch}{1.15}
\begin{tabular}{lccc}
\toprule
Model & $R^2_{\text{EBA}}$ & $R^2_{\text{OPA}}$ & $R^2_{\text{PPA}}$ \\
\midrule
resmlp\_12\_224.fb\_dino              & 0.216 & 0.144 & 0.158 \\
resmlp\_12\_224.fb\_distilled\_in1k   & 0.188 & 0.111 & 0.125 \\
resmlp\_12\_224.fb\_in1k              & 0.181 & 0.107 & 0.122 \\
resmlp\_24\_224.fb\_dino              & 0.228 & 0.148 & 0.161 \\
resmlp\_24\_224.fb\_distilled\_in1k   & 0.183 & 0.109 & 0.123 \\
resmlp\_24\_224.fb\_in1k              & 0.170 & 0.096 & 0.108 \\
resmlp\_36\_224.fb\_distilled\_in1k   & 0.184 & 0.106 & 0.121 \\
resmlp\_36\_224.fb\_in1k              & 0.179 & 0.104 & 0.116 \\
\bottomrule
\end{tabular}
\end{table}

\subsubsection{Vision Transformers (ViT)}
\label{supple:vit}

ViT models use self-attention to capture long-range dependencies in images. We analyzed MAE-pretrained ViT models of base, large, and huge sizes. Table~\ref{tab:vit-summary} shows brain-region alignment scores.

\begin{table}[!t]
\caption{Summary of ViT models pretrained with MAE.}
\label{tab:vit-summary}
\centering
\small
\setlength{\tabcolsep}{6pt}
\renewcommand{\arraystretch}{1.15}
\begin{tabular}{lccc}
\toprule
Model & $R^2_{\text{EBA}}$ & $R^2_{\text{OPA}}$ & $R^2_{\text{PPA}}$ \\
\midrule
vit\_base\_patch16\_224.mae   & 0.217 & 0.151 & 0.163 \\
vit\_large\_patch16\_224.mae  & 0.227 & 0.155 & 0.165 \\
vit\_huge\_patch14\_224.mae   & 0.220 & 0.154 & 0.164 \\
\bottomrule
\end{tabular}
\end{table}

\subsection{Summary}
\label{supple:summary_models}

\begin{itemize}
    \item \textbf{ConvNeXt}: Serves as the primary architecture for controlled analyses, due to consistent backbone, scalable sizes, and pretraining on single datasets. 
    \item \textbf{ResNet, ResMLP, ViT}: Used as cross-architecture checks to verify that key trends are not specific to ConvNeXt.
    \item \textbf{Overall}: Across all architectures, the alignment between model embeddings and neural data demonstrates systematic dependence on pretraining data and model scale, supporting the generality of our findings.
\end{itemize}

\section{Background on Dimensional Analysis}
\label{supple: details of dimension}

\subsection{Fractal Dimension and Correlation Dimension}

The concept of fractal dimension provides a robust framework for quantifying the complexity, irregularity and structural intricacies of datasets across different scales. Unlike traditional Euclidean dimensions, fractal dimensions capture the degree to which a set fills space as the observation scale varies. This is particularly useful in the context of high-dimensional data and manifold learning.

\begin{definition}
\textbf{Fractal Dimension}: The fractal dimension $D_f$ is defined as the scaling exponent that quantifies how the number of self-similar structures in a set changes with the scale of observation. Mathematically, it can be expressed as:
\begin{equation}
D_f = \lim_{\epsilon \to 0} \frac{\log N(\epsilon)}{\log (1/\epsilon)},
\end{equation}
where $N(\epsilon)$ represents the number of $\epsilon$-sized boxes required to cover the set.
A higher fractal dimension indicates a more intricate structure with greater self-similarity across multiple scales.
\end{definition}

\subsubsection{Correlation Dimension}

The correlation dimension is a specific type of fractal dimension that focuses on the spatial distribution properties of a point set, effectively capturing statistical sparsity and clustering behavior. It provides a finer-grained analysis of the dataset structure, especially in cases where data points are non-uniformly distributed.

\begin{definition}
\textbf{Correlation Dimension}: The correlation dimension $D_C$ is defined based on the correlation function $C(\epsilon)$ as:
\begin{equation}
D_C = \lim_{\epsilon \to 0} \frac{\log C(\epsilon)}{\log \epsilon},
\end{equation}
where the correlation function $C(\epsilon)$ is given by:
\begin{equation}
C(\epsilon) = \frac{1}{N(N-1)} \sum_{i=1}^N \sum_{j \neq i} \mathbb{I}(\|x_i - x_j\| < \epsilon).
\end{equation}

\begin{itemize}
    \item $N$ is the number of data points.
    \item $\|x_i - x_j\|$ represents the Euclidean distance between points $x_i$ and $x_j$.
    \item $\mathbb{I}(\cdot)$ is the indicator function, equal to $1$ if the condition inside is true and $0$ otherwise.
\end{itemize}

The correlation dimension effectively estimates the likelihood that two randomly chosen points from the dataset are within a distance $\epsilon$ of each other, thus providing insights into the spatial organization of the data.

\end{definition}

\subsection{Fractal Dimension Estimation Using Maximum Likelihood}

To accurately estimate the fractal dimension, particularly the correlation dimension, we employ the Maximum Likelihood Estimation (MLE) method as proposed by Levina and Bickel \citep{levina2004maximum}. This method leverages distances between neighboring points to infer the intrinsic dimensionality of the underlying manifold.

\subsubsection{Fractal Dimension Estimation via MLE}

The MLE-based estimation method utilizes the $k$-nearest neighbor distances to quantify the dimensionality of a dataset. For a given data point $x_i$, the estimated fractal dimension $\hat{m_k}(x_i)$ at a specific $k$ value is defined as:

\begin{equation}
\hat{m_k}(x_i) = \left[\frac{1}{k-1} \sum_{j=1}^{k-1} \log \frac{T_k(x_i)}{T_j(x_i)}\right]^{-1},
\end{equation}
where:

\begin{itemize}
    \item $T_j(x_i)$ is the Euclidean distance from $x_i$ to its $j^{\text{th}}$ nearest neighbor.
    \item $k$ is the number of nearest neighbors considered for the estimation.
\end{itemize}

A smaller $k$ captures local structural variations, emphasizing finer-scale features, while a larger $k$ reflects broader, more global patterns in the data distribution.

\subsubsection{Averaging Over Data Points}

To obtain a robust estimate of the fractal dimension for the entire dataset, we average the estimated dimensions across all data points:

\begin{equation}
\bar{m_k} = \frac{1}{N} \sum_{i=1}^N \hat{m_k}(x_i),
\end{equation}

where $N$ is the total number of data points. This averaging process helps to mitigate the effects of noise and local variations, producing a stable estimate of the fractal dimension at a specific $k$ value.

\subsubsection{Selection of $k$ and Scale Analysis}

The choice of $k$ significantly impacts the resulting dimension estimate:

\begin{itemize}
    \item \textbf{Small $k$}: Focuses on finer-scale structures, capturing local density variations and small-scale clustering.
    \item \textbf{Large $k$}: Emphasizes broader trends, providing a more smoothed, global perspective of the dataset structure.
\end{itemize}

To achieve a comprehensive understanding of the manifold, it is essential to analyze the fractal dimension estimates across multiple $k$ values. This multi-scale analysis helps in identifying how the estimated dimensionality varies with scale, offering deeper insights into the intrinsic complexity of the dataset.

\section{Supplementary Analysis of Subject Differences}
\label{supple: Subject Differences}

\subsection{Impact of Subject Differences on AI-Brain Alignment}
\label{supple: Subject Differences AI-Brain Alignment}

To further investigate the voxel-wise alignment analysis presented in the main text, we extended the analysis across all eight participants using multiple models. Here, we focus on two key visualizations that highlight the variability and consistency of alignment performance.

\begin{figure}[htbp]
  \centering
  \includegraphics[width=0.95\linewidth]{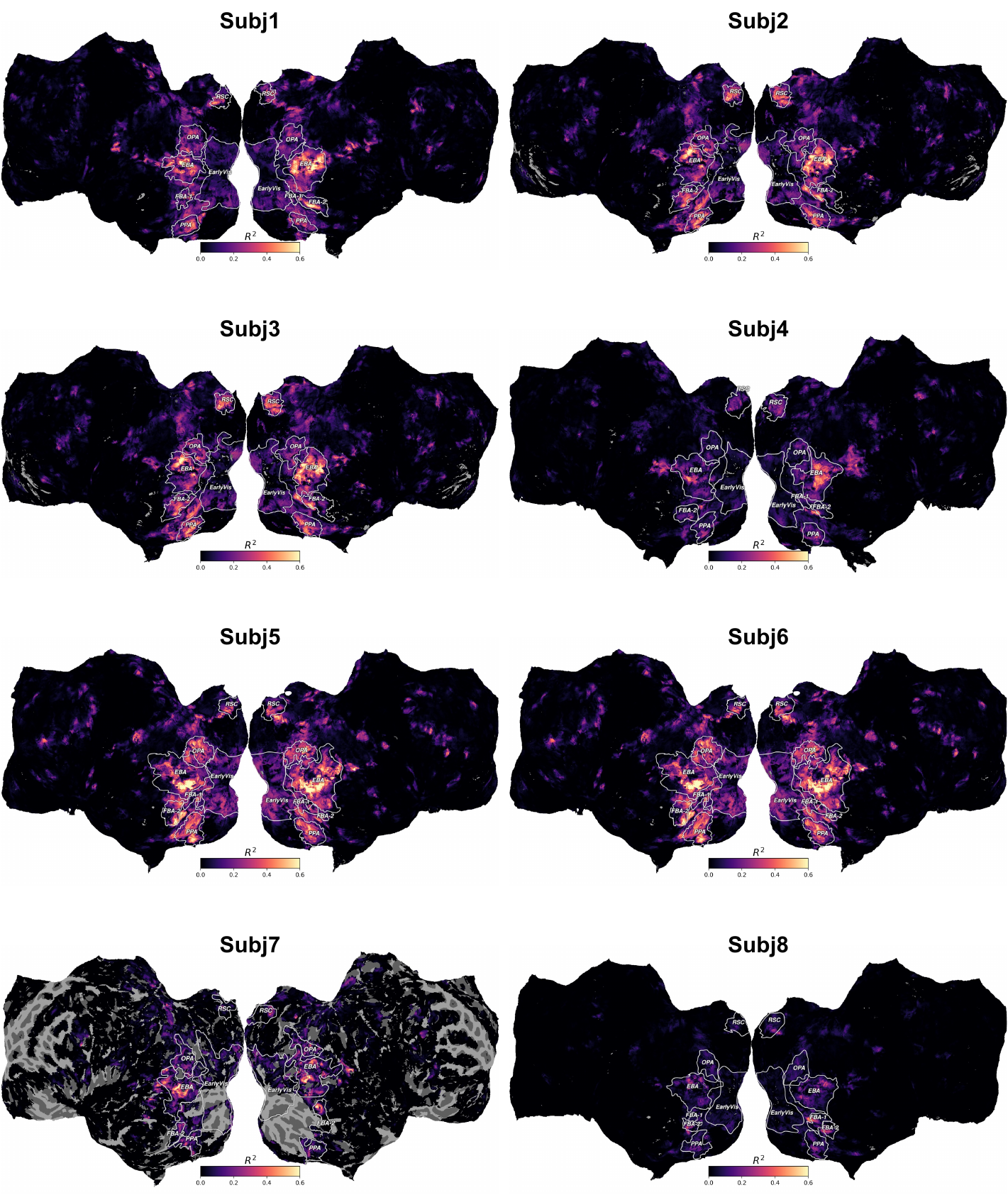}
  \caption{\textbf{Alignment performance across different subjects for the same model (Model index: M50).} This figure illustrates the alignment performance of the same model across different subjects. There is a noticeable variation in alignment effectiveness, with Subjects 5 and 6 exhibiting better alignment, while Subjects 7 and 8 show relatively weaker alignment. This highlights substantial inter-subject variability in alignment performance.}
  \label{fig:alignment_sameModel_diffSubj}
\end{figure}

Figure \ref{fig:alignment_sameModel_diffSubj} illustrates the alignment performance of a selected model across the eight participants. Despite inter-subject variability, the overall spatial distribution of alignment remains relatively consistent across participants. Regions around the EBA consistently show higher alignment, indicating that model-derived representations align more effectively with neural activity in higher-order visual areas. This consistent spatial distribution suggests that the alignment patterns observed in the main analysis are not specific to a single participant but are generalizable across multiple subjects.

\subsection{Impact of Subject Differences on the Correlation between AI-Brain and AI-AI Alignment}
\label{supple:appendix_subjectDiff_AIBrain_AIAI}

We examine whether the correlation between AI-Brain alignment and AI-AI alignment is consistent across individual subjects. For this analysis, we focus on the EBA region and compute, for each subject, the correlation between AI-Brain alignment and AI-AI alignment across models.

\begin{figure}[htbp]
  \centering
  \includegraphics[width=0.98\linewidth]{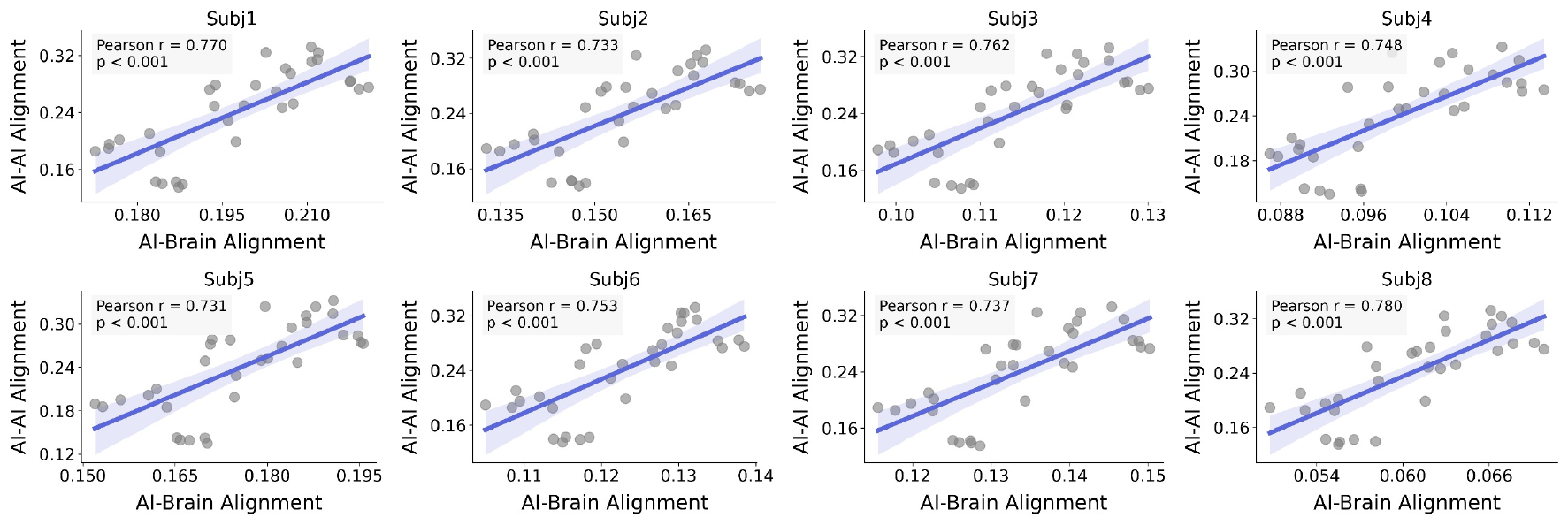}
  \caption{\textbf{Correlation between AI-Brain alignment and AI-AI alignment across subjects in EBA.} Each panel corresponds to a distinct subject. For each subject, we report the Pearson correlation across models. Significant positive correlations are consistently observed, demonstrating that the relationship between AI-AI and AI-Brain alignment is robust to individual differences.}
  \label{fig:appendix_subjectDiff_AIBrain_AIAI}
\end{figure}

As shown in Figure~\ref{fig:appendix_subjectDiff_AIBrain_AIAI}, significant positive correlations are observed for all subjects. The results indicate that models exhibiting stronger alignment with other AI models also tend to better match brain representations, and this relationship is robust across individual differences in neural responses.

\subsection{Impact of Subject Differences on the Correlation between AI-Brain Alignment and Generalization}
\label{supple:appendix_subjectDiff_AIBrain_generalization}

We further investigate whether the relationship between AI-Brain alignment and model generalization performance is affected by subject variability. For each subject in EBA, we compute the correlation between AI-Brain alignment and test performance across models.

\begin{figure}[htbp]
  \centering
  \includegraphics[width=0.98\linewidth]{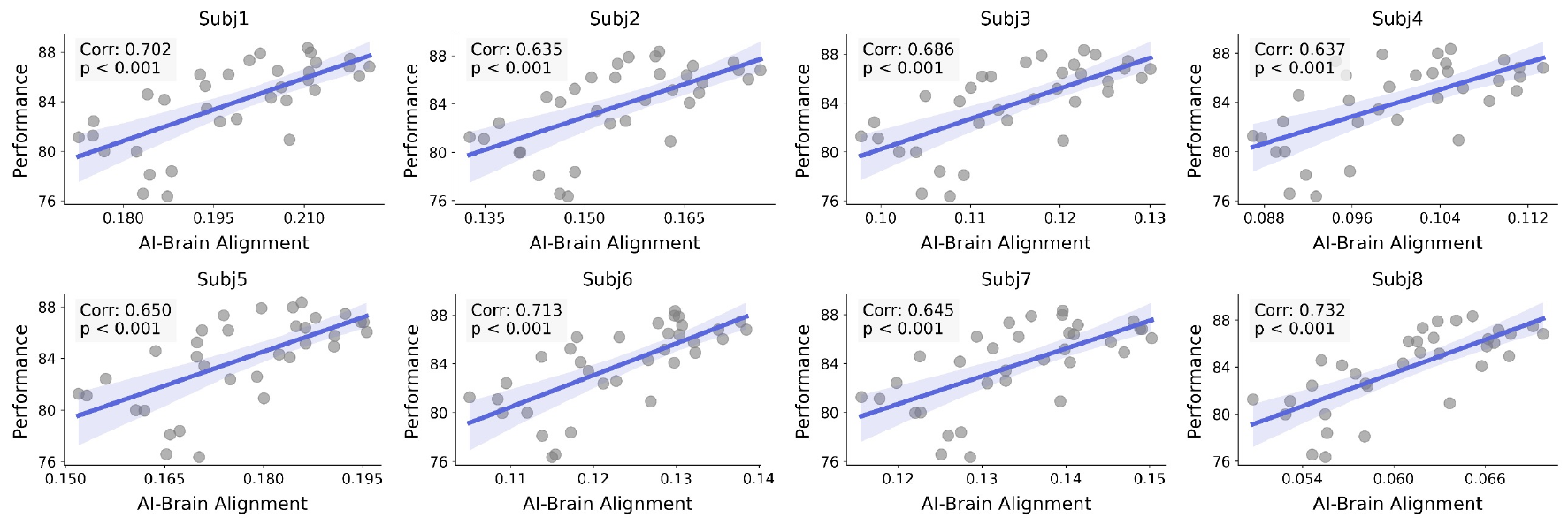}
  \caption{\textbf{Correlation between AI-Brain alignment and generalization performance across subjects in EBA.} Each panel corresponds to a distinct subject. For all subjects, the correlation is positive and statistically significant, indicating that alignment with EBA representations reliably predicts model generalization, independent of individual differences.}
  \label{fig:appendix_subjectDiff_AIBrain_generalization}
\end{figure}

Figure~\ref{fig:appendix_subjectDiff_AIBrain_generalization} shows that all subjects exhibit a significant positive correlation, indicating that models which align better with brain activity in EBA also tend to generalize more effectively. This finding confirms that the predictive power of AI-Brain alignment for generalization is consistent across individual subjects.

\subsection{Impact of Subject Differences on the Correlation between Dimension and AI-Brain Alignment}
\label{supple:appendix_subjectDiff_dimension_AIBrain}

Finally, we assess whether the relationship between representational dimensionality and AI-Brain alignment varies across subjects. For each subject, we compute the correlation between the estimated intrinsic dimensionality of model representations and AI-Brain alignment in EBA.

\begin{figure}[htbp]
  \centering
  \includegraphics[width=0.98\linewidth]{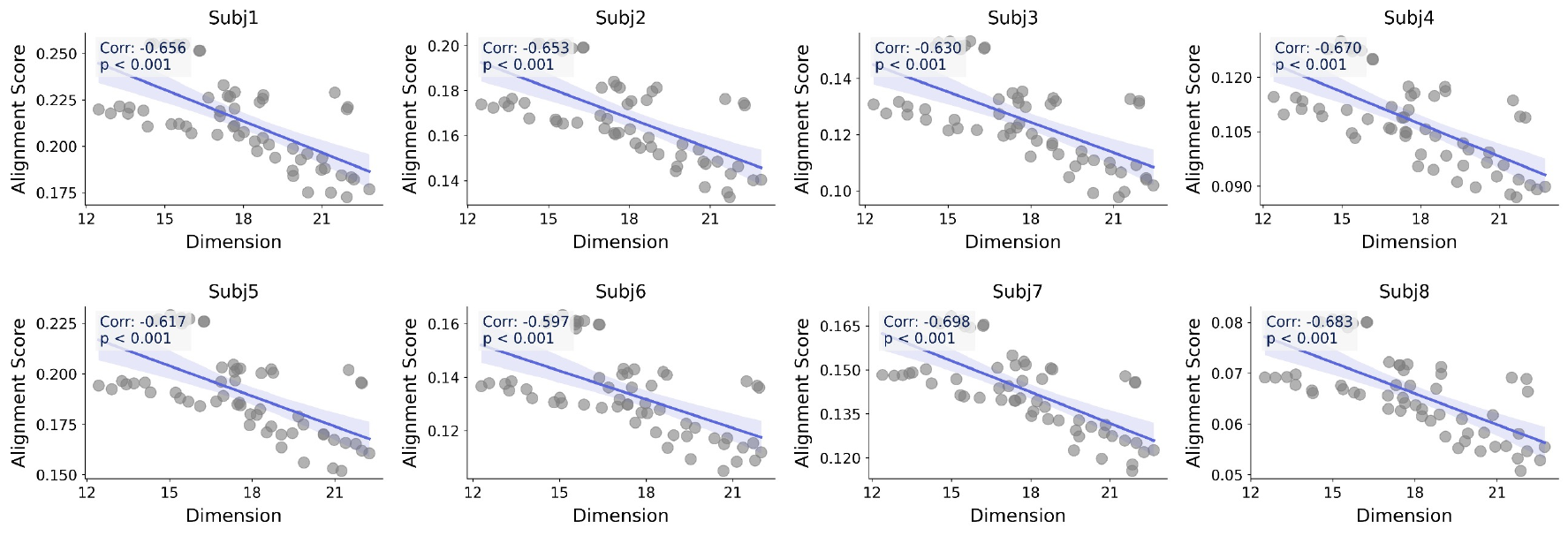}
  \caption{\textbf{Correlation between representational dimensionality and AI-Brain alignment across subjects in EBA.} Each panel corresponds to a distinct subject. Significant positive correlations are observed for all subjects, indicating that the relationship between intrinsic dimensionality and brain alignment is robust to individual differences.}
  \label{fig:appendix_subjectDiff_dimension_AIBrain}
\end{figure}

As depicted in Figure~\ref{fig:appendix_subjectDiff_dimension_AIBrain}, all subjects show significant positive correlations, suggesting that the link between dimensionality and brain alignment is robust to individual variability and reflects a general geometric property of model representations.

\section{Supplementary Analysis of Region Differences}
\label{supple: Region Differences}


In the main text, our analysis primarily focuses on the extrastriate body area (EBA), where AI-Brain alignment shows the strongest and most stable effects. In this section, we extend our analysis to multiple visual regions of interest (ROIs) to examine whether the observed relationships generalize across the visual hierarchy and to assess the influence of regional differences on AI-Brain alignment.

\subsection{Impact of Region Differences on the Correlation between AI-Brain and AI-AI Alignment}
\label{supple:appendix_regionDiff_AIBrain_AIAI}


We first examine whether the correlation between AI-Brain alignment and AI-AI alignment, as reported in the main text, is specific to EBA or persists across other visual regions. For each ROI, we compute the correlation between AI-Brain alignment and AI-AI alignment across models.

\begin{figure}[htbp]
  \centering
  \includegraphics[width=0.98\linewidth]{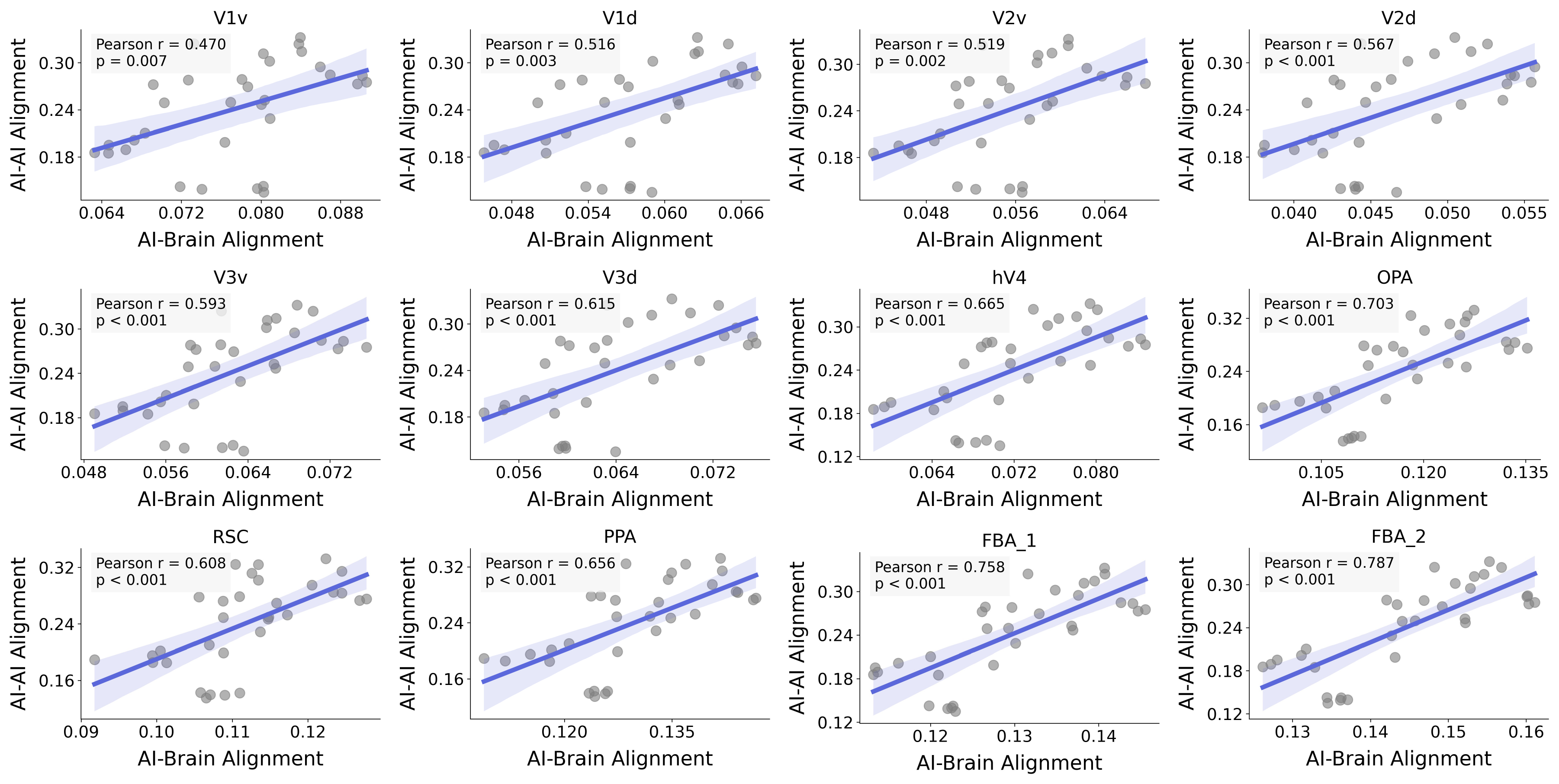}
  \caption{\textbf{Correlation between AI-Brain alignment and AI-AI alignment across different visual regions.} Each panel corresponds to a distinct region of interest. For each region, we report the Pearson correlation between AI-Brain alignment and AI-AI alignment across models. Significant positive correlations are consistently observed across regions, indicating that the relationship between AI-Brain and AI-AI alignment is robust to regional differences.}
  \label{fig:appendix_regionDiff_AIBrain_AIAI}
\end{figure}

As shown in Figure~\ref{fig:appendix_regionDiff_AIBrain_AIAI}, a significant positive correlation is consistently observed across a wide range of visual regions, including early (V1, V2), intermediate (V3, hV4), and higher-level visual areas. Although the absolute strength of the correlation varies across regions, the overall trend remains stable, indicating that models that better align with other AI models also tend to exhibit stronger alignment with brain representations, independent of the specific cortical region considered. This result demonstrates that the coupling between AI-AI and AI-Brain alignment is not restricted to a single functional area but reflects a more general representational property.

\subsection{Impact of Region Differences on the Correlation between AI-Brain Alignment and Generalization}
\label{supple:appendix_regionDiff_AIBrain_generalization}


Next, we analyze how AI-Brain alignment in different visual regions relates to model generalization performance. For each ROI, we compute the correlation between AI-Brain alignment and test performance across models.

\begin{figure}[htbp]
  \centering
  \includegraphics[width=0.98\linewidth]{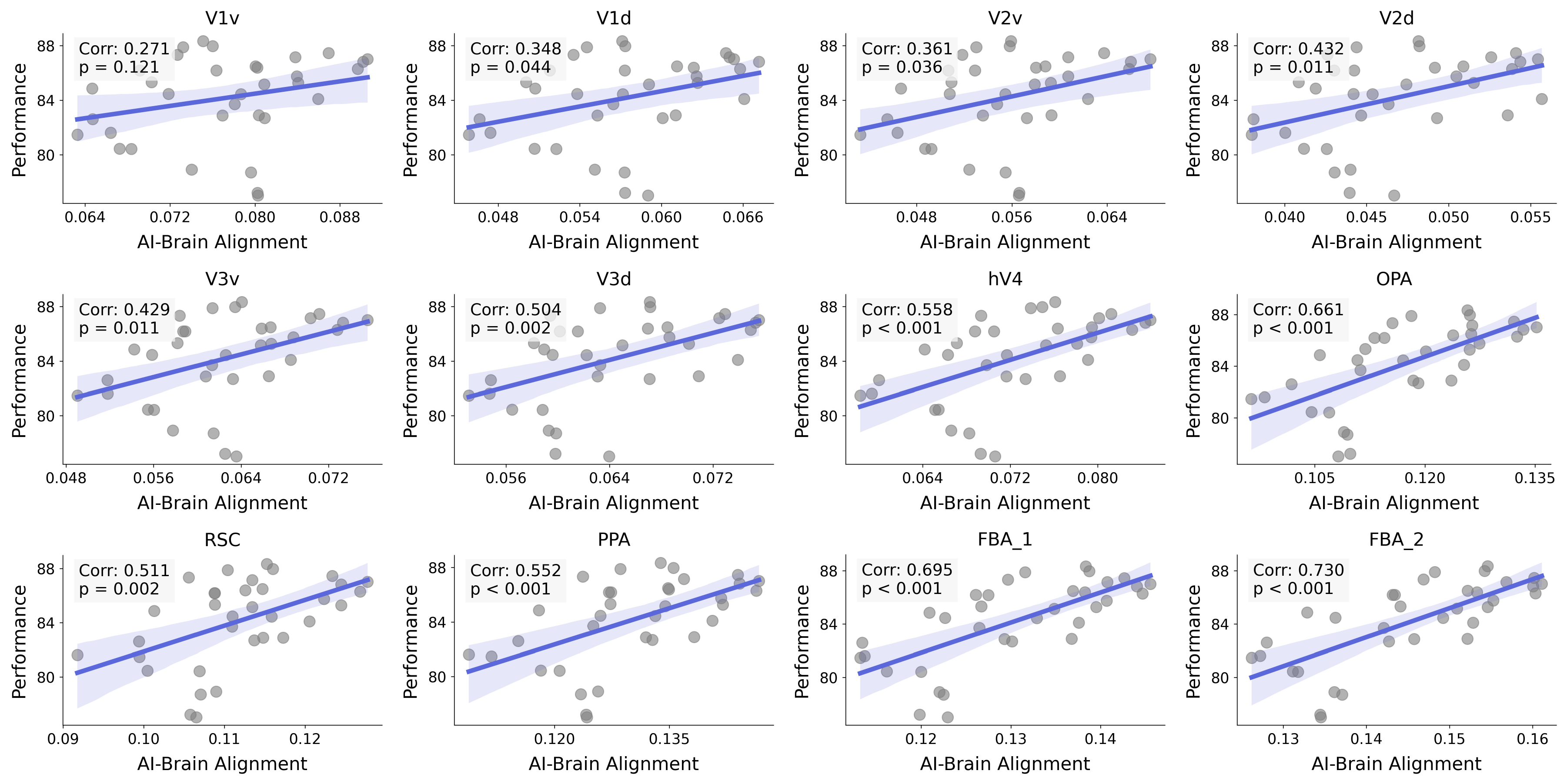}
  \caption{\textbf{Correlation between AI-Brain alignment and generalization performance across visual regions.} Each panel shows the relationship between AI-Brain alignment and test performance for a specific ROI. The correlation is weak and non-significant in early visual areas (e.g., V1) but increases progressively along the visual hierarchy, becoming strongest and statistically significant in higher-level visual regions.}
  \label{fig:appendix_regionDiff_AIBrain_generalization}
\end{figure}

Figure~\ref{fig:appendix_regionDiff_AIBrain_generalization} reveals a clear regional dependence along the visual hierarchy. In early visual cortex (e.g., V1), the correlation between AI-Brain alignment and generalization performance is weak and not statistically significant. However, as visual information progresses through the hierarchy from V1 to V2 and hV4, the correlation strength gradually increases. In higher-level visual regions, the correlation becomes both statistically significant and substantially stronger.

These results suggest that alignment with higher-order visual representations is more predictive of a model's ability to generalize, whereas alignment with early sensory representations alone is insufficient. This hierarchical pattern is consistent with the functional roles of these regions and supports the interpretation that AI-Brain alignment captures behaviorally relevant representational structure more effectively in higher visual areas.

\subsection{Impact of Region Differences on the Correlation between Dimension and AI-Brain Alignment}
\label{supple:appendix_regionDiff_dimension_AIBrain}


We further investigate whether the relationship between representational dimensionality and AI-Brain alignment depends on the cortical region considered. For each ROI, we compute the correlation between the estimated intrinsic dimensionality of model representations and AI-Brain alignment.

\begin{figure}[htbp]
  \centering
  \includegraphics[width=0.98\linewidth]{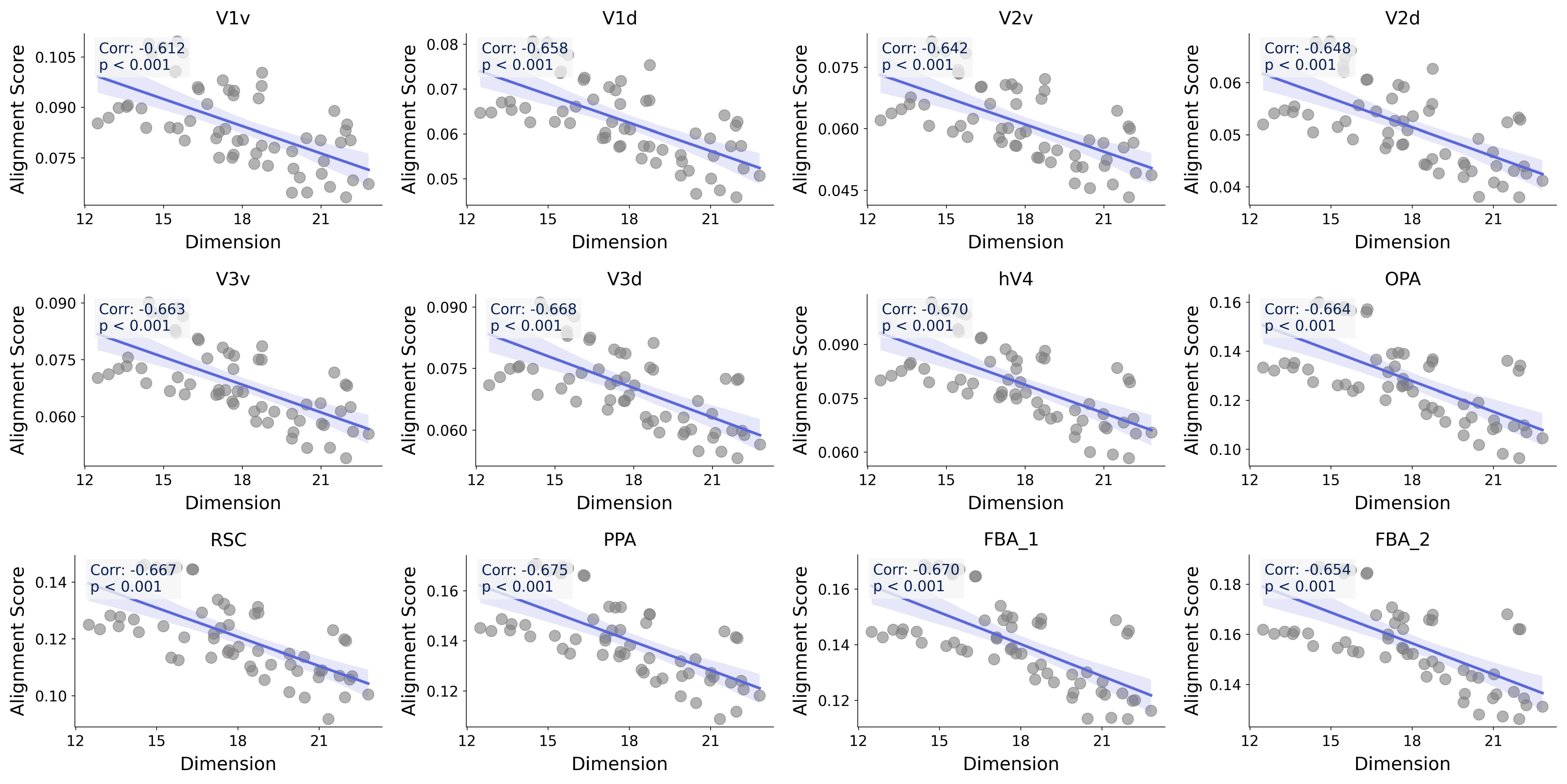}
  \caption{\textbf{Correlation between representational dimensionality and AI-Brain alignment across visual regions.} For each ROI, we report the correlation between intrinsic dimensionality and AI-Brain alignment across models. Significant correlations are observed in all regions, with no clear monotonic increase along the visual hierarchy.}
  \label{fig:appendix_regionDiff_dimension_AIBrain}
\end{figure}

As shown in Figure~\ref{fig:appendix_regionDiff_dimension_AIBrain}, dimensionality is significantly correlated with AI-Brain alignment across all examined regions. Unlike the analysis with generalization performance, we do not observe a systematic increase in correlation strength from early to higher-level visual areas. Instead, the dimensionality-alignment relationship appears relatively stable across regions.

This result suggests that the link between representational dimensionality and AI-Brain alignment reflects a general geometric property of representations rather than a region-specific effect tied to higher-level visual processing.

\section{Supplementary Analysis of Dimensionality Estimators}
\label{supple:estimator_differences}

In the main text, we emphasize the importance of local representational structure by analyzing dimensionality at different neighborhood scales. Here, we further assess the robustness of these findings with respect to the choice of dimensionality estimator.

\subsection{Impact of Estimator Differences on Scale Analysis}
\label{supple:appendix_estimator_normalAnalysis}



We first evaluate whether the scale-dependent relationships reported in the main text depend on a specific dimensionality estimation method. In addition to the estimator used in the main analysis, we consider two alternative estimators, MOM and MADA. For each estimator and neighborhood scale, we compute the correlation between estimated dimensionality and AI-AI alignment, AI-Brain alignment, and generalization performance using all samples.

\begin{figure}[!t]
  \centering
  \includegraphics[width=0.98\linewidth]{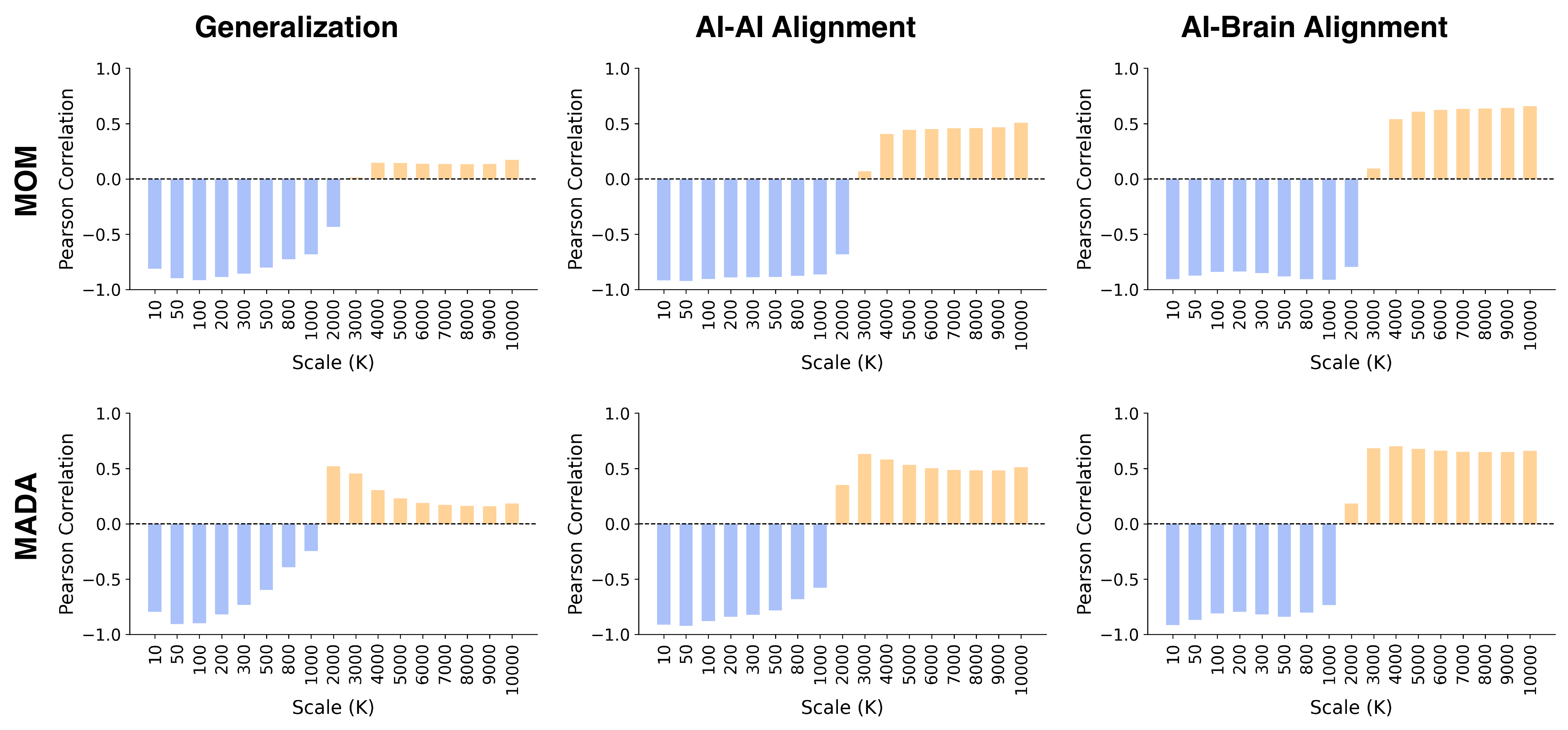}
  \caption{\textbf{Effect of dimensionality estimator choice on scale-dependent correlations.} Correlations between estimated dimensionality and AI-AI alignment, AI-Brain alignment, and generalization performance are shown for different estimators and neighborhood sizes. The qualitative trends are consistent across estimators, with the strongest effects occurring at small neighborhood scales.}
  \label{fig:appendix_estimator_normalAnalysis}
\end{figure}

Figure~\ref{fig:appendix_estimator_normalAnalysis} shows that the overall trends are highly consistent across estimators. At larger neighborhood scales, correlations tend to weaken and may even change sign, while at smaller scales, dimensionality exhibits strong and significant correlations with alignment and generalization. These results indicate that the observed scale-dependent effects are not artifacts of a particular estimator but reflect intrinsic properties of the representations.

\subsection{Impact of Estimator Differences on Local Structure Analysis}
\label{supple:appendix_estimator_localAnalysis}


Finally, we repeat the local structure analysis using alternative dimensionality estimators. Following the procedure in the main text, we estimate dimensionality based on local neighborhoods of 1,000 nearest neighbors and analyze its relationship with AI-AI alignment, AI-Brain alignment, and generalization performance.

\begin{figure}[!t]
  \centering
  \includegraphics[width=0.98\linewidth]{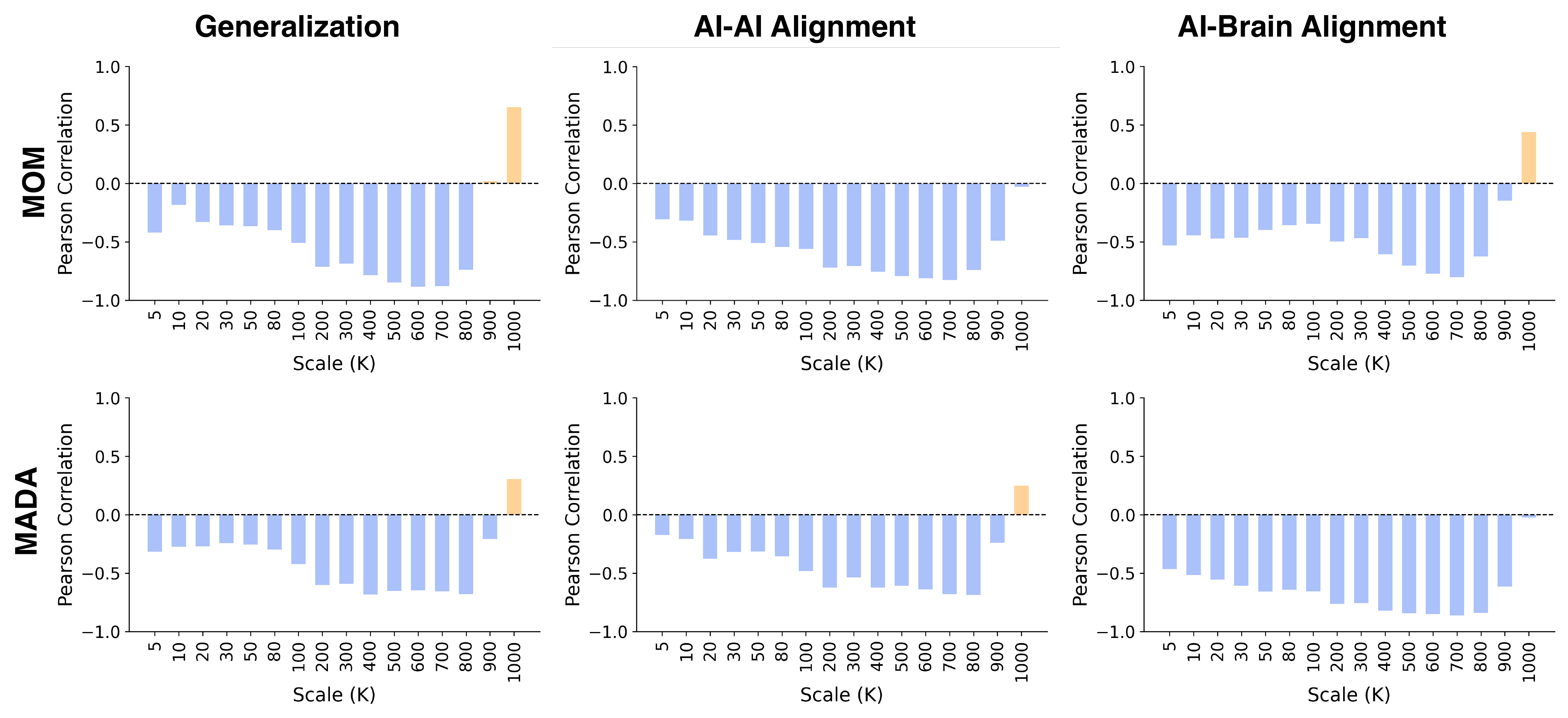}
  \caption{\textbf{Robustness of local structure analysis across dimensionality estimators.} Correlations between local intrinsic dimensionality and AI-AI alignment, AI-Brain alignment, and generalization performance are shown for different estimators. Significant correlations are consistently observed, supporting the robustness of the local structure findings.}
  \label{fig:appendix_estimator_localAnalysis}
\end{figure}

As shown in Figure~\ref{fig:appendix_estimator_localAnalysis}, all estimators yield significant correlations across a range of neighborhood sizes. The consistency across estimators further confirms that the importance of local representational geometry is robust and not driven by estimator-specific biases.

\section{Supplementary Analysis of Architecture Differences}
\label{supple:architecture_differences}



\begin{figure}[htbp]
  \centering
  \includegraphics[width=0.98\linewidth]{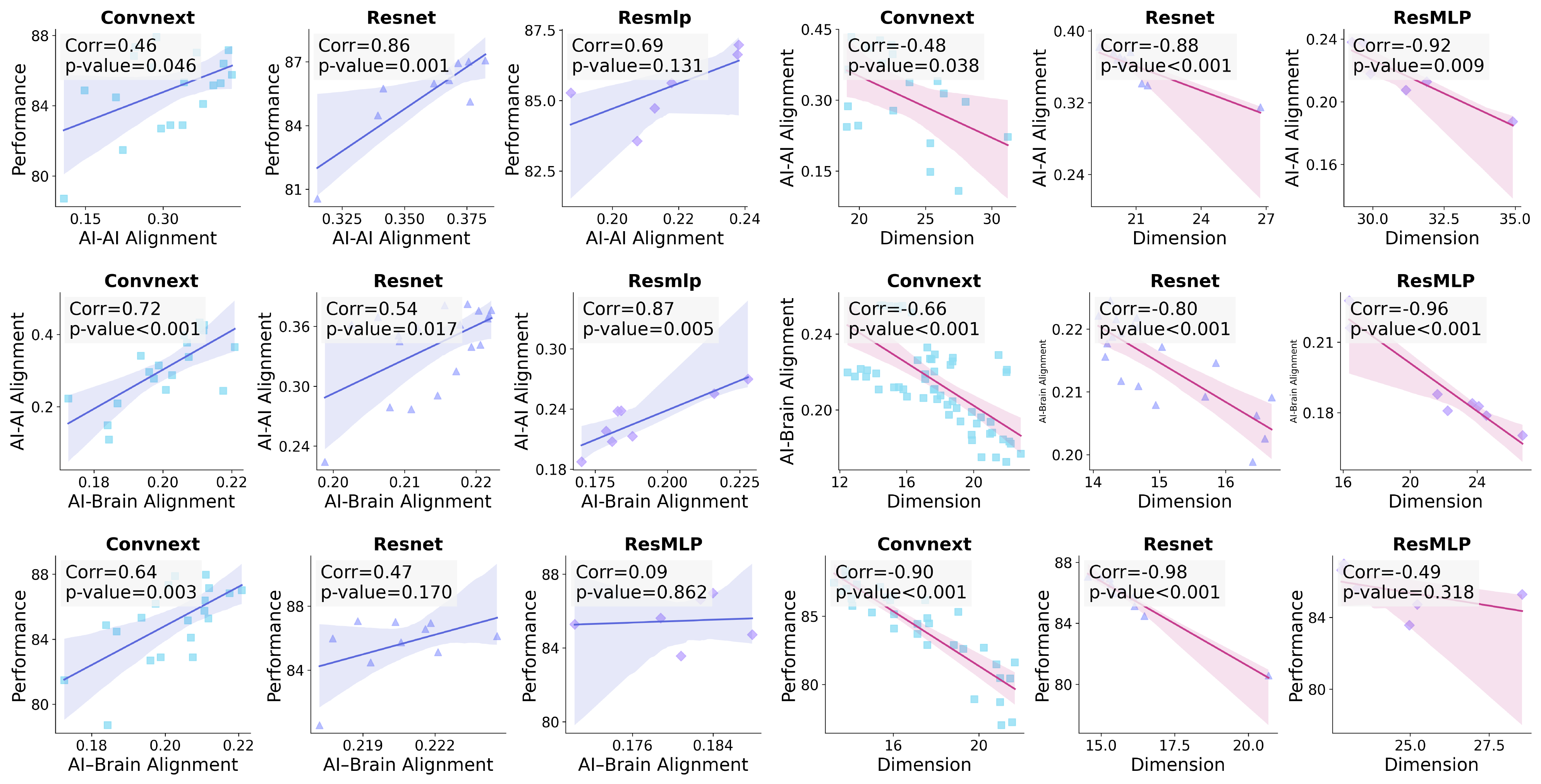}
  \caption{
  \textbf{Single-architecture analyses across ResNet, ResMLP and ConvNeXt models.} 
  Relationships between inter-model alignment, model-to-fMRI alignment, generalization performance, and local embedding dimensionality are shown. Overall, trends observed in ConvNeXt are largely preserved across ResNet and ResMLP, supporting cross-architecture generality. Dimensionality consistently correlates with AI-AI alignment, AI-Brain alignment, and generalization performance, highlighting its importance as a geometric descriptor of model embeddings.
  }
  \label{fig:appendix_singleModel}
\end{figure}

To validate whether the findings reported in the main text generalize across different model families, we conducted additional analyses on ResNet and ResMLP architectures. Specifically, we examined inter-model alignment, model-to-fMRI alignment, generalization performance, and local embedding dimensionality, as shown in Figure~\ref{fig:appendix_singleModel}. 

Overall, the results are consistent across architectures. Notably, the relationship between AI-Brain Alignment and generalization performance is less pronounced in ResNet and ResMLP models. This is likely because models with higher generalization performance exhibit very similar alignment with fMRI signals (e.g., EBA $R^2$ ranging from 0.217-0.224 for ResNet, 0.175-0.285 for ResMLP), reducing variability across models. 

Interestingly, ConvNeXt models achieve the highest generalization performance among all architectures. Despite ResNet and ResMLP being structurally distinct, they still exhibit significant correlations between AI-Brain Alignment, AI-AI Alignment, and generalization performance. In particular, ResNet models show that better generalization is associated with higher alignment to fMRI responses and closer embeddings to ConvNeXt models. These observations suggest that representational convergence may occur across different architectures.

Furthermore, local embedding dimensionality, as a geometric property, is significantly correlated with AI-AI Alignment, AI-Brain Alignment, and generalization performance across architectures. This underscores the importance of dimensionality as a robust metric for characterizing model representations.

\end{document}